\documentclass[final,5p,times,twocolumn]{elsarticle}
\pdfoutput=1

\usepackage{mathtools} % 数式の縦方向のスペース
\usepackage{amsmath}
\usepackage{array}
\usepackage{setspace}
\usepackage{arydshln}
\usepackage{siunitx}

\usepackage{comment} % コメントアウト
\usepackage[usetype1]{uline--} 

\usepackage{bm}
\usepackage{txfonts}
\usepackage{multirow}
\usepackage{threeparttable} % 表の注釈
\usepackage{diagbox} % 斜線

\usepackage{algpseudocode} % 疑似コード
\usepackage{algorithm}     % これも疑似コード
\usepackage{here}
\usepackage{midpage} % 図を中心へ配置
\usepackage[subrefformat=parens]{subcaption} % サブキャプション

\usepackage{enumitem}

\setlist[enumerate]{leftmargin=*}
\setlist[itemize]{leftmargin=*}

\begin{document}

\begin{frontmatter}

\title{Cardiomyopathy Diagnosis Model from Endomyocardial Biopsy Specimens:\\Appropriate Feature Space and Class Boundary in Small Sample Size Data}

\author{Masaya Mori\fnref{label1}}
\author{Yuto Omae\fnref{label1}}
\author{Yutaka Koyama\fnref{label2}}
\author{Kazuyuki Hara\fnref{label1}}
\author{Jun Toyotani\fnref{label1}}
\author{Yasuo Okumura\fnref{label3}}
\author{Hiroyuki Hao\fnref{label2}\corref{cor1}}

\cortext[cor1]{Corresponding author. E-mail address: hao.hiroyuki@nihon-u.ac.jp}

\affiliation[label1]{organization={College of Industrial Technology, Nihon University},
            addressline={1-2-1 Izumi},
            city={Narashino},
            state={Chiba},
            postcode={275-8575},
            country={Japan}}
\affiliation[label2]{organization={Department of Pathology and Microbiology, Nihon University School of Medicine},
            addressline={30-1 Oyaguchi Kamicho},
            city={Itabashi},
            state={Tokyo},
            postcode={173-8610},
            country={Japan}}
\affiliation[label3]{organization={Department of Medicine, Nihon University School of Medicine},
            addressline={30-1 Oyaguchi Kamicho},
            city={Itabashi},
            state={Tokyo},
            postcode={173-8610},
            country={Japan}}

%% Abstract
\begin{abstract}
As the number of patients with heart failure increases, machine learning (ML) has garnered attention in cardiomyopathy diagnosis, driven by the shortage of pathologists.
However, endomyocardial biopsy specimens are often small sample size and require techniques such as feature extraction and dimensionality reduction.
This study aims to determine whether texture features are effective for feature extraction in the pathological diagnosis of cardiomyopathy.
Furthermore, model designs that contribute toward improving generalization performance are examined 
by applying feature selection (FS) and dimensional compression (DC) to several ML models.
The obtained results were verified by visualizing the inter-class distribution differences and conducting statistical hypothesis testing based on texture features.
Additionally, they were evaluated using predictive performance across different model designs 
with varying combinations of FS and DC (applied or not) and decision boundaries.
The obtained results confirmed that texture features may be effective for the pathological diagnosis of cardiomyopathy.
Moreover, when the ratio of features to the sample size is high, a multi-step process involving FS and DC 
improved the generalization performance, with the linear kernel support vector machine achieving the best results.
This process was demonstrated to be potentially effective for models with reduced complexity, 
regardless of whether the decision boundaries were linear, curved, perpendicular, or parallel to the axes.
These findings are expected to facilitate the development of an effective cardiomyopathy diagnostic model for its rapid adoption in medical practice.
\end{abstract}

%% Keywords
\begin{keyword}
Cardiomyopathy \sep Endomyocardial biopsy \sep Pathology image analysis \sep Machine learning \sep Texture analysis \sep Dimensionality reduction \sep Low sample size


\end{keyword}

\end{frontmatter}


% ▼▼▼▼▼▼▼▼▼▼▼▼▼▼▼▼▼▼▼▼▼▼▼▼▼▼▼▼▼▼▼▼▼▼▼▼
% ------------------------------------------------- 本文 -------------------------------------------------
% ▼▼▼▼▼▼▼▼▼▼▼▼▼▼▼▼▼▼▼▼▼▼▼▼▼▼▼▼▼▼▼▼▼▼▼▼

\section{Introduction}
With the aging of society, heart failure is increasing worldwide. 
Various physiological and imaging examinations, such as echocardiography, cardiac computed tomography, 
and cardiac magnetic resonance imaging are used to determine the causes of heart failure. 
An endomyocardial biopsy is the only in vivo method for obtaining histopathological information about the myocardium~\cite{ishibashi2017significance}. 
It is useful for discriminating between primary cardiomyopathies, such as 
hypertrophic, dilated, restrictive, and arrhythmogenic right ventricular, 
as well as secondary cardiomyopathies, such as cardiac amyloidosis, cardiac sarcoidosis, lymphocytic myocarditis, 
giant cell myocarditis, and Fabry disease~\cite{leone20122011, cooper2007role}.
However, a worldwide shortage of pathologists who can make such a diagnosis has emerged. 
Machine learning (ML) has recently been used for pathological diagnoses~\cite{nirschl2018deep, pallua2020future}.

If ML could be used to enumerate differential diseases based on image analysis and 
automatically calculate the likelihood of each disease, it would be useful for pathological diagnosis. 
It may also prevent misdiagnosis owing to a lack of expert knowledge.
Realizing this objective requires the development of an ML model capable of making pathological diagnoses 
based on histopathological information (myocardial cell diameter, myocardial cell morphology, nuclear diameter, 
nuclear morphology, presence and frequency of cellular infiltration, types of infiltrating cells, 
myocardial cell arrangement, fibrosis, and definitive diagnosis) obtained from myocardial biopsy or 
autopsy specimens of patients with a history of cardiac disease.

\begin{figure*}[t]
  \centering
  \includegraphics[width=\textwidth]{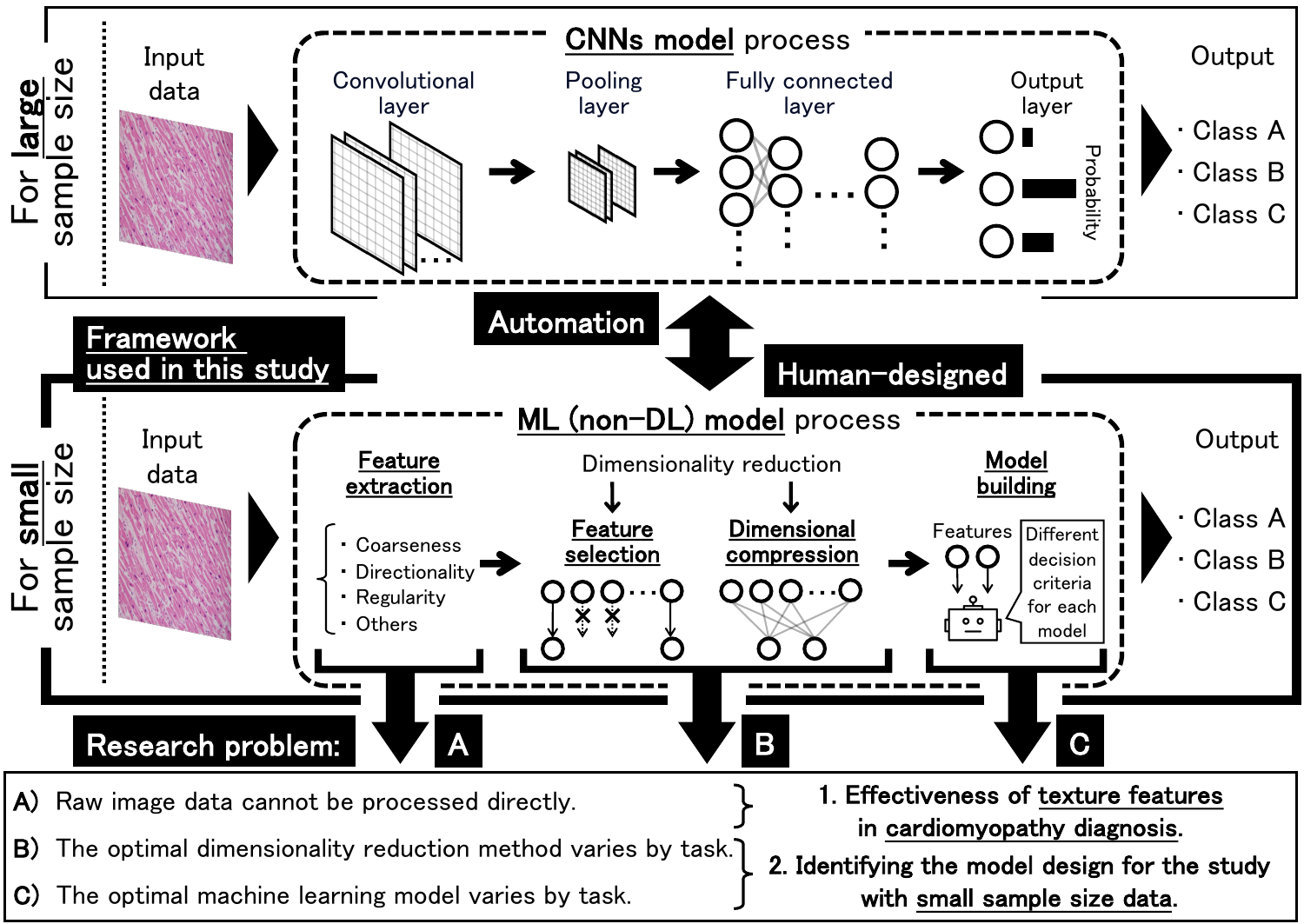}
  \caption{Graphical introduction.}
  \label{fig:GA_intro}
\end{figure*}

Convolutional neural networks (CNNs) are widely used to predict disease states using medical images~\cite{cai2020review, li2023medical}.
One reason for this is that, in contrast to traditional ML methods, 
CNNs automate three processes that would typically require manual intervention 
(Figure~\ref{fig:GA_intro}: ML (non-deep learning (DL)) model process)~\cite{liu2018feature}: 
(1) feature extraction, which quantifies the visual information inherent in the input image (such as color, brightness, patterns, textures, shapes, and object scale),
(2) dimensionality reduction, which aims to improve analysis efficiency and prevent overfitting by selecting useful features for predicting and eliminating irrelevant or noisy features 
(this involves feature selection (FS) to select features deemed useful for recognizing the target, 
and dimensional compression (DC) to transform high-dimensional feature spaces into lower-dimensional ones), and
(3) model building, which maximizes both the goodness of fit on training data and the generalization performance on unseen data 
(both performances are collectively referred to as predictive performance in this study).

%%%%%%%%%  図：イントロのGA  %%%%%%%%%%
%%%%%%%%%%%%%%%%%%%%%%%%%%%%%%%%%%%%

The main layers in CNNs include convolutional, pooling, fully connected, and output (Figure~\ref{fig:GA_intro}: CNNs model process).
In the convolutional layer, convolution operations using kernel filters are performed on the input, generating feature maps that emphasize the important image features for prediction.
In the pooling layer, the spatial local regions within the obtained feature maps are integrated, consolidating the information from these regions.
By alternating these two layers, useful visual information for recognizing the target is extracted from the pixel array of the input image.
Thereafter, the fully connected layer takes the one-dimensional feature map as the input and performs transformations on the features by connecting all the nodes from the previous and subsequent layers. 
Typically, the number of nodes in each layer decreases as the network progresses from shallow to deeper layers.
By repeating this process, the feature map input can be represented as a lower-dimensional feature vector containing useful features for recognizing the target.
Finally, the feature vector obtained in the fully connected layer is fed into the output layer, where, 
in regression tasks, the continuous predicted value is the output, whereas in classification tasks, the probabilities for each class are the outputs.
Therefore, CNNs automate the processes necessary for prediction~\cite{ergun2016early, lokesh2022cnn}.
Consequently, CNNs eliminate the need to design handcrafted features based on deep domain expertise, 
which is typically required in non-DL methods, as well as dimensionality reduction of the feature space.
Moreover, CNNs have the potential to extract features that may not be visible to humans, 
and can thus be expected to achieve a diagnostic performance comparable to, or even surpassing, 
that of expert clinicians~\cite{han2018deep, nagpal2019development}.

However, to achieve a high predictive performance with CNNs, 
a vast amount of image data is required to sufficiently capture the diversity of the training data. 
Therefore, CNNs applied in the medical field often target X-ray, CT, and MRI images
~\cite{khalid2020comparative, wu2022chest, sharifrazi2022cnn, aromiwura2023artificial}. 
This is because these tests are non-invasive, widely used in clinical settings, and it is easier to collect high-quality images.
In contrast, endomyocardial biopsy is a highly invasive procedure with associated risks, which necessitates careful examination~\cite{from2011current}. 
Consequently, the frequency of these tests is relatively low, 
making it difficult to collect a sufficiently diverse set of histopathological images of the myocardium
~\cite{tong2017predicting, dooley2018prediction}. 
Histopathological images of the myocardium often include small sample sizes with insufficient diversity.
When constructing predictive models using small sample size data, 
it has been reported that the risk of overfitting increases when the ratio of features to the sample size or 
the complexity of ML models is high~\cite{vabalas2019machine}. 
Here, model complexity refers to the extent to which a model can capture data patterns, 
which is determined by the structure of the algorithm and the number and values of the hyperparameters.
Therefore, it is necessary to reduce the dimensionality of the dataset prior to training or adopt ML models with lower complexity. 
In conclusion, for the pathological diagnosis of cardiomyopathy 
using small sample histopathological images of the myocardium, 
ML models with lower complexity that handle relatively low-dimensional data, 
such as non-DL models, are considered more suitable than complex CNNs that require high-dimensional input data.

Moreover, from the perspective of interpretability of the diagnostic rationale, non-DL ML models are considered more suitable. 
In deep image recognition models based on convolutions, 
methods that apply class activation mapping (CAM) are widely used to visualize the rationale behind predictions
~\cite{porumb2020convolutional, yildiz2021classification, pikulkaew2023enhancing}.
CAM visually emphasizes the areas within the input image that contribute significantly to the prediction results~\cite{zhou2016learning}.
One advantage of this approach is that the reasoning behind the model can be confirmed visually, 
allowing even those unfamiliar with ML to understand it intuitively. 
This makes it easy to compare the reasoning of the model with the judgment rationale of experts in the relevant field. 
However, as CAM provides qualitative reasoning, it is challenging to identify the specific conditions that contribute 
to the prediction in non-structured images, which lack distinct shapes~\cite{amato2024explainable}.
Therefore, in the case of cardiomyopathy pathology images, 
it is difficult to interpret the specific reasons for the predictions from the CAM, 
such as color, surface roughness, regularity, and directionality.
In such cases, the application of statistical methods, as used in non-DL models for feature extraction and selection, 
is considered effective for interpreting the specific grounds for predictions.

However, to achieve a pathological diagnosis of cardiomyopathy using a non-DL ML model, 
it is necessary to consider (1) the extraction of image features 
from histopathological images of the myocardium and (2) model design, which is robust to small sample size data.

First, the extraction of image features from cardiomyopathy pathology images is described 
(Figure~\ref{fig:GA_intro}: Research problem A). 
Raw images contain a wide variety of visual information, such as color, shape, patterns, texture, and composition. 
This visual information is interrelated and numerically represented at the pixel level. 
Therefore, when analyzing the visual information of an entire image, raw images are interpreted as high-dimensional data, 
represented by the product of the height, width, and depth. 
These data include not only the visual information necessary for prediction 
(e.g., information about the animal itself in animal classification) 
but also irrelevant visual information (e.g., background information in animal classification). 
These irrelevant details increase the risk of the ML model generating incorrect decision criteria, which can reduce its generalization performance. 
Furthermore, in the case of high-dimensional data, the sample size required for learning is enormous, making it difficult to ensure sufficient data density. 
A decrease in data density is one factor that can prevent the model from appropriately learning important features for prediction. 
Therefore, to improve the generalization performance, 
it is essential to extract the relevant visual information that contributes toward solving the problem from high-dimensional data 
and summarize it into a lower-dimensional feature vector~\cite{kumar2020detailed}.

In DL, a process is performed to summarize diverse visual information embedded 
in image data into numerical representations using several parameters across multiple layers.
In contrast, non-DL models generally lack such transformation mechanisms, 
making feature extraction necessary to summarize the visual information of image data into meaningful features.
Texture information, such as surface patterns and regularities of histopathological images of the myocardium, 
are generally considered useful for differentiating cardiomyopathies~\cite{shirani2000morphology, cianci2024morphological}. 
This suggests that texture features are particularly suitable for feature extraction 
from histopathological images of the myocardium.
Texture features are widely employed in the construction of diagnostic models using medical images in ML, 
and their effectiveness has been widely reported~\cite{castellano2004texture, chowdhary2020segmentation}.
However, to the best of our knowledge, few studies have been conducted to evaluate the effectiveness of texture features 
from histopathological images of the myocardium for the pathological diagnosis of cardiomyopathies using ML.
Thus, it is crucial to verify whether texture features from histopathological images of the myocardium 
are effective in predicting the pathological diagnosis of cardiomyopathies.

Next, the design of models robust to small sample size data is discussed (Figure~\ref{fig:GA_intro}: Research problem B).
Data with small sample sizes often present various challenges, 
such as a reduction in the data density within the feature space~\cite{phan2012multiscale} and domain shift~\cite{guan2021domain}.
Collectively, these issues contribute to the risk of overfitting.
Therefore, achieving high generalization performance with small sample size data requires model designs that demonstrate robust performance against these challenges.

When constructing ML models using small sample size data, the use of a reduced-dimensional feature space is recommended~\cite{mori2022prediction}. 
This approach increases data density by reducing dimensionality, which is expected to enhance generalization performance. 
However, the degree to which the dimensionality should be reduced may vary depending on the specific task. 
For instance, a higher-dimensional feature space allows the incorporation of more information regarding target prediction, 
potentially improving the prediction performance. 
Conversely, the inclusion of irrelevant information may also increase, 
resulting in the generation of decision criteria based on meaningless data, which could cause overfitting and reduce the generalization performance. 
Therefore, it is essential to carefully determine an acceptable level of dimensionality.

In addition, it is important to consider not only the dimensionality of the feature space 
but also the complexity of the ML model (Figure~\ref{fig:GA_intro}: Research problem C). 
A model with controlled complexity can avoid overfitting to the training data~\cite{aliferis2024overfitting}. 
Consequently, even with a small sample size data, high generalization performance on unseen data can be expected. 
However, within this context, decision boundaries such as straight lines, curves, 
or boundaries perpendicular or parallel to the axes may exist, and the suitability of the decision boundary depends on the specific task.
In general, models that use linear decision boundaries tend to exhibit simpler internal structures, which helps suppress overfitting. 
However, achieving high prediction performance may be difficult~\cite{aliferis2024overfitting}. 
Conversely, models with nonlinear decision boundaries exhibit relatively complex internal structures, 
which can result in a higher goodness of fit, but may suffer from lower generalization performance~\cite{aliferis2024overfitting}. 
To construct a classification model that approximates the optimal prediction performance, 
it is necessary to carefully determine an acceptable level of model complexity.

Based on the above, this study investigates two aspects: the effectiveness of texture features 
from histopathological images of the myocardium for pathological diagnosis of cardiomyopathies, 
and the design of models suitable for pathological diagnosis of cardiomyopathies using small sample size data. 
As reported in previous research, we constructed a preliminary ML model 
based on a single model design and performed a simple investigation of the potential of texture features 
for pathological diagnosis of three types of cardiomyopathies~\cite{mori2024potential}. 
The obtained results indicated that texture features clearly represent differences in myocardial conditions. 
Based on this, this study constructs a three-class classification model based on texture features 
and multiple model designs.
Through visualization and statistical analysis of texture features, 
and performance evaluation of the classification models, 
a more rigorous validation of the effectiveness of the texture features and the optimal model design was conducted. 
The clarification of these aspects will contribute to the pathological diagnosis of cardiomyopathies 
based on high diagnostic performance in environments where large sample sizes cannot be collected. 
Moreover, it is anticipated that, in the future, rapid and accurate diagnoses enable timely treatment for patients.

The remainder of this paper is organized as follows:
Section~\ref{sec:ML (non-DL) methods} provides an overview of the ML methods used in this study.
Section~\ref{sec:Experiment} describes the experimental procedures, details of the datasets, and the parameters employed in the experiments.
Section~\ref{sec:Results and discussion} presents the experimental results and discusses their implications.
Section~\ref{sec:Conclusions} concludes the paper with a summary of the study, its limitations, and future prospects.

\section{Machine learning (non-deep learning) methods} \label{sec:ML (non-DL) methods}
\subsection{Texture feature extraction} \label{subsec:Texture feature extraction}
In this study, the following texture analysis methods were employed: 
first-order statistics (FOS), gray level difference statistics (GLDS), gray level co-occurrence matrix (GLCM), gray level run length matrix (GLRLM), 
angular distribution function via Fourier power spectrum (ADF), and radial distribution function via Fourier power spectrum (RDF).
FOS refers to statistical measures derived from a distribution in which the gray level values of image pixels 
are plotted on the X-axis, and the corresponding occurrence probabilities are plotted on the Y-axis.
GLDS represents statistical measures derived from distributions in which the gray level differences 
between a given pixel and its neighboring pixel, 
at a specified direction and distance, are plotted on the X-axis, 
with the occurrence probabilities of these differences plotted on the Y-axis. 
These distributions are calculated for the horizontal, vertical, and diagonal directions, 
and the resulting statistics are averaged across all directions.
GLCM is a matrix that aggregates the occurrence probabilities of pairs of gray level values 
in pixels adjacent to a given pixel at specified distances and directions.
GLRLM is a matrix that aggregates the occurrence probabilities of consecutive runs 
of pixels with the same grey-level value in a specified direction.
ADF refers to the distribution derived from the power spectrum image obtained via a two-dimensional Fourier transform, 
where the angular directions from the center of the spectrum are plotted on the X-axis, 
and the intensity of the frequency components along these directions is plotted on the Y-axis.
RDF is a distribution in which the distance from the center of the power spectrum image 
obtained via a two-dimensional Fourier transform is plotted on the X-axis, 
and the intensity of the frequency components at that distance is plotted on the Y-axis.

Texture features were calculated from various statistical measures derived from the distributions and matrices obtained using texture analysis methods. 
Table~\ref{table:texture features} lists the texture analysis methods and their corresponding statistical measures. 
For FOS, GLDS, ADF, and RDF, seven statistical measures were adopted: mean, contrast, variance, skewness, kurtosis, energy, and entropy.
The mean quantifies the center of mass of the probability distribution. 
The value decreases as the distribution shifts to the left and increases as it shifts to the right. 
The contrast is an indicator that quantifies the magnitude of the probability distribution, excluding the sign.
Regardless of the sign, it increased as the probability distribution moved further from 0.
Variance quantifies the extent to which the data deviate from the mean of the probability distribution. 
The value is smaller when the data are concentrated around the mean, and larger when they are spread out. 
Skewness quantifies the asymmetry of a probability distribution compared to a normal distribution. 
If the distribution is skewed to the left with a longer right tail, the value is positive; if it is skewed to the right with a longer left tail, the value is negative. 
In addition, the magnitude of skewness increases as the degree of asymmetry increases, resulting in larger positive or negative values.
Kurtosis is a measure that quantifies the sharpness of the peak and the heaviness of the tails of a probability distribution in comparison with a normal distribution. 
When the distribution peaks have heavier tails than a normal distribution, 
kurtosis has a positive value; when the distribution is flatter with lighter tails, it has a negative value. 
Furthermore, the stronger this tendency, the larger the value is in the positive or negative direction.
Energy quantifies the concentration of the probability distribution. 
The value is higher when the distribution is concentrated at specific values, and lower when it is spread out over a wider range. 
Entropy quantifies the closeness of a distribution to a uniform distribution. 
The value increases as the distribution approaches uniformity, and decreases as the distribution concentrates around specific values.

For GLCM, six statistical measures were adopted: contrast, correlation, joint energy, joint entropy, inverse difference moment (IDM), and inverse variance.
Correlation is a measure that quantifies the linear dependence of shade changes between adjacent pixels.
Strong linear dependence between shade changes in adjacent pixels 
(e.g. when bright pixels are surrounded by similar pixels and the shade changes in a step-like manner), the value is close to 1.
However, when there is a weak linear dependence (e.g. when bright pixels are surrounded by dark pixels), the value is close to 0.
Joint energy quantifies the degree of continuity of specific shade patterns between adjacent pixels.
The larger the value, the more frequently a specific shade pattern appears between adjacent pixels, while the smaller the value, 
the more evenly distributed the various shade patterns are.
Joint entropy quantifies the diversity and irregularity of shade patterns between adjacent pixels.
The larger the value, the more diverse and irregular the shade patterns are between adjacent pixels, while the smaller the value, 
the more biased towards a specific shade pattern.
IDM is an index that quantifies the uniformity and smoothness of an image.
The larger the value, the more uniform is the image, with pixels of equal tone values appearing successively; conversely, 
the smaller the value, the more uneven is the image, with pixels of high- and low-tone values appearing in succession.
Inverse variance is a measure that quantifies the variation in changes in grayscale values between adjacent pixels. 
The larger the value, the more frequently small differences in grayscale values occur between adjacent pixels, indicating less variation. 
Conversely, the smaller the value, the larger the differences in grayscale values between adjacent pixels, indicating greater variation.

For GLRLM, five statistical measures were adopted: short run emphasis (SRE), 
long run emphasis (LRE), gray level non-uniformity (GLN), run length non-uniformity (RLN), and run percentage (RP).
SRE is an index that quantifies the fineness of an image.
The higher the value, the shorter the run length in the image (the finer the image).
LRE is an index that quantifies the coarseness of an image.
The higher the value, the longer is the run length in the image (the coarser the image).
GLN is an index that quantifies the uniformity of the distribution of grayscale values.
The larger the value, the more frequently a particular gray value appears compared with other gray values (the distribution of gray values is uneven).
However, the smaller the value, the more even the gray values appear (the distribution of gray values is even).
RLN quantifies the evenness of the distribution of run lengths.
The larger the value, the more frequently a particular run length appears compared with other run lengths (the distribution of run lengths is uneven).
In contrast, the more even the run lengths (the more uniform the run length distribution), the smaller the value.
RP quantifies the coarseness or fineness of a texture.
The shorter the number of runs (the more detailed the image), the larger the value; conversely, the longer the runs (the coarser the image), the smaller the value.

In this study, a Python package was developed to analyze the texture information of images and compute their statistical features. 
By utilizing this package, a total of 39 texture features, as shown in Table~\ref{table:texture features}, were calculated. 
The features related to GLCM and GLRLM were implemented using the Python package Pyradiomics (version: 3.1.0)~\cite{van2017computational}.

Because the scale of each texture feature is different, considering the importance of each feature equally, 
it is necessary to make the features dimensionless through standardization.
For small sample size datasets, each feature is highly likely to be non-normally distributed. 
Furthermore, when applying standardization, robustness to outliers is necessary.
Therefore, robust standardization, a standardization method 
that is robust to outliers and is not affected by the shape of the distribution, was adopted here.
Scaling through robust standardization was performed as follows: 
\begin{align}
  z(x) = \frac{x - \tilde{x}}{\mathrm{IQR}}.
\end{align}
where $x$ is an element of the texture feature, $\tilde{x}$ is the median of the texture feature, and $\mathrm{IQR}$ is the interquartile range of the texture feature.

%%%%%%%%%  表：テクスチャ特徴量  %%%%%%%%%%
\begin{table*}[t]
  \centering
  
  \begin{threeparttable}[h] % 注釈
  \caption{Texture analysis methods and their corresponding statistical measures ($\bigcirc$ indicates applied).}
  \label{table:texture features}

  \begin{tabular*}{\textwidth}{@{\extracolsep{\fill}}lcccccc}
    \hline
    \multirow{2}{*}{Statistical measures$^{1}$} & \multicolumn{6}{c}{Texture analysis methods$^{2}$} \\
                     & FOS & GLDS & GLCM & GLRLM & ADF & RDF \\  
    \hline
    Mean             & $\bigcirc$ & $\bigcirc$ &\textemdash{}&\textemdash{}& $\bigcirc$ & $\bigcirc$ \\
    Contrast         & $\bigcirc$ & $\bigcirc$ & $\bigcirc$  &\textemdash{}& $\bigcirc$ & $\bigcirc$ \\
    Variance         & $\bigcirc$ & $\bigcirc$ &\textemdash{}&\textemdash{}& $\bigcirc$ & $\bigcirc$ \\
    Skewness         & $\bigcirc$ & $\bigcirc$ &\textemdash{}&\textemdash{}& $\bigcirc$ & $\bigcirc$ \\
    Kurtosis         & $\bigcirc$ & $\bigcirc$ &\textemdash{}&\textemdash{}& $\bigcirc$ & $\bigcirc$ \\
    Energy           & $\bigcirc$ & $\bigcirc$ &\textemdash{}&\textemdash{}& $\bigcirc$ & $\bigcirc$ \\
    Entropy          & $\bigcirc$ & $\bigcirc$ &\textemdash{}&\textemdash{}& $\bigcirc$ & $\bigcirc$ \\
    \hdashline
    Correlation      &\textemdash{}&\textemdash{}& $\bigcirc$ &\textemdash{}&\textemdash{}&\textemdash{}\\
    Joint energy     &\textemdash{}&\textemdash{}& $\bigcirc$ &\textemdash{}&\textemdash{}&\textemdash{}\\
    Joint entropy    &\textemdash{}&\textemdash{}& $\bigcirc$ &\textemdash{}&\textemdash{}&\textemdash{}\\
    IDM              &\textemdash{}&\textemdash{}& $\bigcirc$ &\textemdash{}&\textemdash{}&\textemdash{}\\
    Inverse variance &\textemdash{}&\textemdash{}& $\bigcirc$ &\textemdash{}&\textemdash{}&\textemdash{}\\
    \hdashline
    SRE              &\textemdash{}&\textemdash{}&\textemdash{}& $\bigcirc$ &\textemdash{}&\textemdash{}\\
    LRE              &\textemdash{}&\textemdash{}&\textemdash{}& $\bigcirc$ &\textemdash{}&\textemdash{}\\
    GLN              &\textemdash{}&\textemdash{}&\textemdash{}& $\bigcirc$ &\textemdash{}&\textemdash{}\\
    RLN              &\textemdash{}&\textemdash{}&\textemdash{}& $\bigcirc$ &\textemdash{}&\textemdash{}\\
    RP               &\textemdash{}&\textemdash{}&\textemdash{}& $\bigcirc$ &\textemdash{}&\textemdash{}\\
    \hline
  \end{tabular*}

  \footnotesize{
  \begin{tablenotes}
    \item[$1$] IDM stands for inverse difference moment.
    SRE stands for short run emphasis.
    LRE stands for long run emphasis.
    GLN stands for gray level non-uniformity.
    RLN stands for run length non-uniformity.
    RP stands for run percentage.
    \item[$2$] FOS stands for first-order statistics.
    GLDS stands for gray level difference statistics.
    GLCM stands for gray level co-occurrence matrix.
    GLRLM stands for gray level run length matrix.
    ADF stands for angular distribution function via fourier power spectrum.
    RDF stands for radial distribution function via fourier power spectrum.
  \end{tablenotes}
  }
  
  \end{threeparttable}
\end{table*}
%%%%%%%%%%%%%%%%%%%%%%%%%%%%%%%%%%%%

\subsection{Feature selection} \label{subsec:Feature selection}
FS is the process of selecting features that are useful for prediction and excluding features that may be noisy or not useful. 
This is performed using an algorithm suited to the characteristics of the dataset and the problem being addressed.
In this study, the within-class variance between-class variance ratio was adopted as the FS method.
In this method, the total sum of the Euclidean distances between the average vectors of all samples and 
the average vectors of each class were divided by the sum of the variances of each class, which was considered the final evaluation value.
Features with sufficiently separated inter-class distances and small within-class variances are desirable; therefore, a higher evaluation value is preferable.
This method has the advantage of easily selecting features that can clearly distinguish different classes of classification problems, 
because it considers the distance between classes and the variance of each class simultaneously.
In addition, because the separation relationship between the classes is simple, 
it is possible to obtain a feature space that is not prone to overfitting, even in the case of small sample size data.
Another advantage is that the selected features are easy to interpret.
However, there is a drawback in that it is not possible to accurately evaluate the spatial structure seen in anomaly detection, where a specific class is concentrated at a single point in the feature space and the other classes are scattered around it.
Such a spatial structure is expected to contribute to an improved prediction performance; however, 
this also increases the risk of overfitting, making it difficult to interpret the selected features.
Therefore, these features were excluded from the selection in this study.

In this study, the within-class variance between-class variance ratio was applied individually to each texture feature. 
The FS evaluation value is calculated using one dimension at a time.
This was done to reduce the search range for FS and reduce the calculation and time costs.
For example, when evaluating a one-dimensional feature from 39 feature types, ${}_{39} \mathrm{C}_1 = 39$ different evaluations are required. 
When evaluating a four-dimensional feature space, ${}_{39} \mathrm{C}_4 = 82,251$ different evaluations were required. 
When evaluating a seven-dimensional feature space ${}_{39} \mathrm{C}_7 = 15,380,937$ different evaluations are required.
It can be observed that the higher the dimensionality of the evaluated space, 
the wider the search range, making it more difficult to obtain an answer that approximates the optimal solution in a realistic amount of time.

On the other hand, for a one-dimensional evaluation, the existence of feature values that can clearly classify multiple classes in one dimension is a prerequisite.
However, there is a possibility that such feature values do not exist.
Therefore, simplification of feature value selection was attempted by dividing the multiclass classification problem into multiple two-class classification problems.
For example, in a three-class classification problem ($\mathrm{C_{a}}$, $\mathrm{C_{b}}$, $\mathrm{C_{c}}$) 
was divided into a combination of multiple two-class classification problems 
($\mathrm{C_{a}}$, $\mathrm{C_{b}}$), ($\mathrm{C_{a}}$, $\mathrm{C_{c}}$), ($\mathrm{C_{b}}$, $\mathrm{C_{c}}$).
Thereafter, the search for useful features for the two-class classification problem was conducted for each of these, 
and the union of the obtained features, excluding duplicates, was extracted as a feature useful for three-class classification.
That is, if $n$ useful features are selected for each two-class classification problem, 
then $3n$ features are selected for the entire three-class classification problem.
If there are $d$ overlapping features among them, then $3n - d$ features are selected after excluding them.

\subsection{Dimensional compression} \label{subsec:Dimensional compression}
DC is a method for converting data into a low-dimensional space 
while minimizing the loss of information regarding the spatial structure in a multidimensional feature space.
In this study, the supervised DC method Fisher's linear discriminant analysis (LDA) was adopted.

LDA is a method for performing a linear transformation that maximizes the between-class variance and 
minimizes the within-class variance based on the training data~\cite{fisher1936use}.
In this method, by maximizing the feature values that contribute to class identification, 
each class is expected to be clearly separated, even after transformation to a low-dimensional space.
However, the dimensionality of the transformed space depends on the number of classes $c$,
and the dimensionality of the transformed space is constrained to $c-1$ or less~\cite{brown2000discriminant}.
Therefore, the number of dimensions that can be selected is only one dimension for two-class classification 
and only one or two dimensions for three-class classification.
In this study, two-dimensional compression was adopted, 
which considers the interaction of feature values because it targets three-class classification.

The LDA implementation used the Python package scikit-learn (version: 1.2.2)~\cite{sklearn_api}.

\subsection{Classification models and evaluation function} \label{subsec:Classification models}
For small sample size data, a prediction model with 
limited complexity is used to avoid overfitting~\cite{heikonen2023modeling}.
However, the degree of complexity depends not only on the sample size and number of features 
but also on the domain, such as the difficulty of solving a problem~\cite{aliferis2024overfitting}.
Therefore, it is not always optimal to choose a simple model, and it is believed that a certain degree of complexity 
is necessary at the decision boundary to achieve a high prediction performance.
In this study, to verify this, a support vector machine (SVM) was adopted 
as a simpler model compared to other classification models.
In this case, a linear kernel (LK) and radial basis function kernel (RBF) were adopted as the kernel functions.
In addition, a decision tree (DT) was adopted as the model with a decision boundary 
that is perpendicular or horizontal to the axis based on a tree structure algorithm.
This was performed to verify the appropriate complexity of the decision boundary for myocardial pathology diagnosis 
by comparing the evaluation values based on the linearity and nonlinearity of the decision boundaries.
These were implemented using the Python package scikit-learn (version: 1.2.2)~\cite{sklearn_api}.

The prediction performance of each classification model is evaluated based on the macro-average F1-score.
The F1-score is an evaluation metric for two-class classification 
that uses the harmonic mean of the precision and recall as its evaluation value.
Precision is a metric that indicates the proportion of samples that are predicted to be positive or actually positive.
Recall is a measure of the proportion of samples 
that were correctly predicted as positive among the samples that were actually positive.
The F1-score enables the prediction performance of each class to be reflected equally,
even in the case of unbalanced data with bias in the samples for each class in two-class classification.
The macro-average F1-score is an indicator that can be extended to multiple classes, 
and is effective when all classes are evaluated equally.
This evaluation value takes a value in the range 0--1, 
and it can be interpreted such that the closer it is to 1, the better the prediction performance.

\subsection{Cross-validation and hyperparameter optimization} \label{subsec:Cross-validation and hyperparameter optimization}
\begin{figure*}[t]
  \centering
  \includegraphics[width=\textwidth]{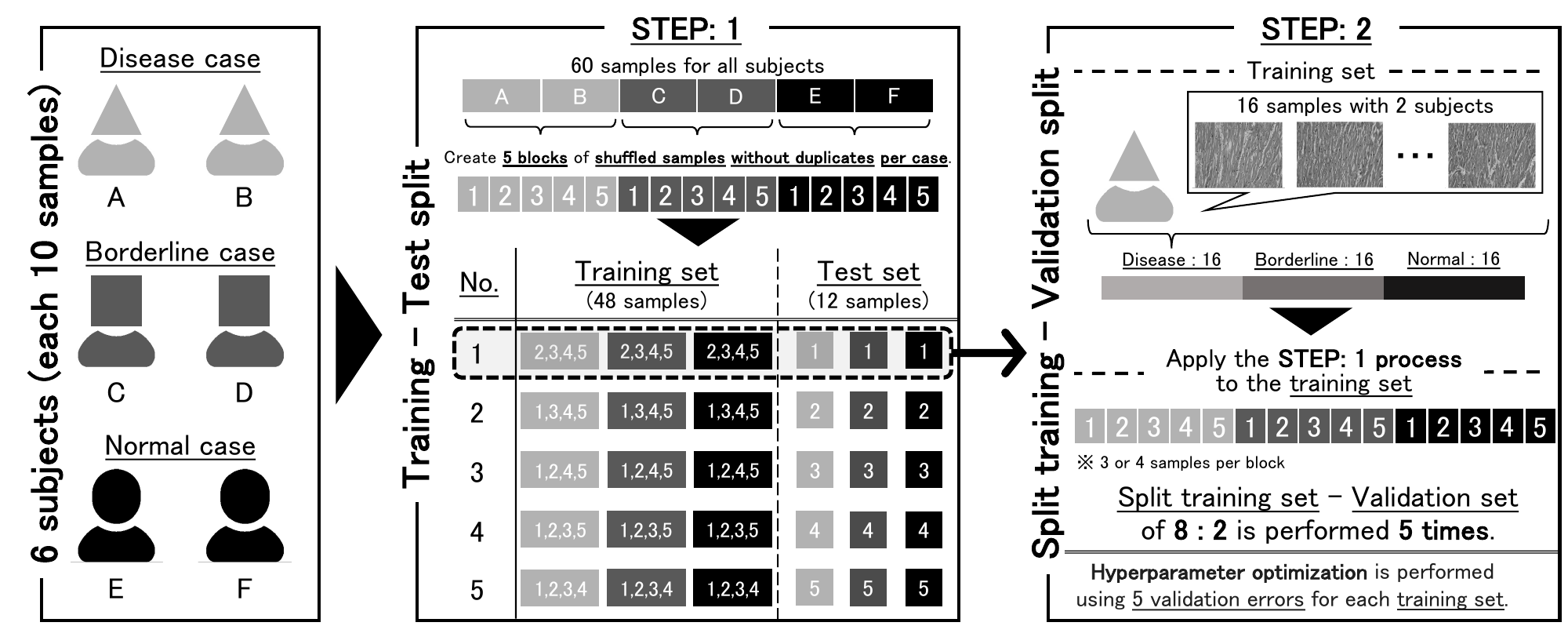}
  \caption{Overview of stratified $K$-fold cross-validation with nested structure for $K=5$.}
  \label{fig:Cross-validation}
\end{figure*}

Cross-validation is a method for statistically evaluating a model's generalization performance 
by repeatedly constructing and evaluating it using a subset of the original dataset as validation or test data, 
while the remaining data are used for training.
In this study, stratified $K$-fold cross-validation was adopted as a cross-validation method.
This method creates a subset from the original dataset, such that the sample ratio of each class is equal.
For example, consider a dataset with 20 samples of disease cases, 20 samples of borderline cases, 
and 20 samples of normal cases, which is then divided into training data and test data 
(Figure~\ref{fig:Cross-validation}: STEP 1).
The sample size ratio was set to training:test = 8:2. 
Five subsets were created, each containing four disease cases, 
four borderline cases, and four normal cases, with no overlapping samples between the subsets. 
The training set was then formed by combining four of these subsets 
(16 disease cases, 16 borderline cases, and 16 normal cases), while the remaining subset 
(four disease cases, four borderline cases, and four normal cases) was allocated to the test set. 
Finally, this subset replacement was performed five times. 
The generalization performance of the model was statistically evaluated based on the evaluation values of the five subsets.
The same procedure applies to the division of the split training and validation set (Figure~\ref{fig:Cross-validation}: STEP 2).

Hyperparameter optimization based on validation error minimization was performed 
when cross validating the split training and validation sets.
Hyperparameter optimization is the process of adjusting hyperparameters 
that influences the complexity of a ML model based on an optimization method.
In this study, the hyperparameters of texture analysis methods 
and the hyperparameters of the ML models were the targets of optimization.
In this case, Bayesian Optimization was adopted as the optimization method.
The hyperparameters that are the target of optimization are shown in Table~\ref{table:hyperparameters}.
For hyperparameters not listed here, the default values of the Python packages used in each method were adopted.

%%%%%%%%%  図：交差検証の解説図  %%%%%%%%%%
%%%%%%%%%%%%%%%%%%%%%%%%%%%%%%%%%%%%

The bin width in Table~\ref{table:hyperparameters} is a parameter 
that determines the discretization of grayscale image tone values.
For example, if the bin width is set to 64, a conversion process is applied, in which 0 becomes black and 63 becomes white.
The distance in Table~\ref{table:hyperparameters} is a parameter 
that determines the distance between the pixels to be compared when comparing the tone values between the pixels.
For example, if the distance is set to 1, a comparison is made between the pixel and the next pixel. 
If the distance is set to 2, then a comparison is made between pixels and two pixels away.
The infinitesimal angle in Table~\ref{table:hyperparameters} is a parameter 
which determines the smoothness of the angular distribution.
The smaller this value, the smoother the angular distribution and the more detailed the trend that can be captured. 
However, its disadvantage is that it requires a longer processing time.
The infinitesimal radius in Table~\ref{table:hyperparameters} is a parameter 
which determines the smoothness of the radial distribution.
The smaller this value, the smoother the radial distribution and the more detailed the trend that can be captured. 
However, its disadvantage is that it requires a longer processing time.
The number of FS-selected features in Table~\ref{table:hyperparameters} is a parameter 
that determines the number of features selected in the FS.
The same applies to number of FS-selected features with DC.
The smaller this value is, the fewer features are selected; thus, there is a greater chance of excluding features which are important for prediction.
On the other hand, the larger this value is, the more features are selected; 
thus, there is a greater chance of selecting features that could be noisy and unnecessary for prediction.
The cost parameter in Table~\ref{table:hyperparameters} is a parameter 
determines the extent to which misclassification is allowed when there is a mixture of samples.
The smaller the value, the more misclassification is allowed, and the larger the value, the less misclassification is allowed.
The $\gamma$ parameter in Table~\ref{table:hyperparameters} is a parameter that determines the complexity of the decision boundary.
The smaller the value, the simpler the decision boundary, and the larger the value, the more complex the decision boundary.
The criterion in Table~\ref{table:hyperparameters} is a hyperparameter that determines how to split nodes in a DT.
``gini'' is the Gini impurity, ``entropy'' is the information gain based on Shannon entropy, 
and ``log loss'' is the cross-entropy loss.
All these metrics measure the uncertainty and/or mixture of the class distribution.
The max depth in Table~\ref{table:hyperparameters} is a hyperparameter that determines the maximum depth of the DT.
The smaller this value, the simpler is the DT structure, which helps reduce the risk of overfitting.
However, the larger this value, the more complex the DT structure becomes, 
leading to a higher goodness of fit on the training data but also an increased risk of reduced generalization performance.

Stratified $K$-fold cross-validation was implemented using the Python package scikit-learn (version: 1.2.2)~\cite{sklearn_api}.
Bayesian Optimization was implemented using the Python package optuna (version: 3.5.0)~\cite{optuna_2019}.

%%%%%%%%%  表：ハイパーパラメーター一覧  %%%%%%%%%%
\begin{table*}[t]
  \centering
  \begin{threeparttable}[h] % 注釈
  \caption{Hyperparameters to be optimized and their search ranges.}
  \label{table:hyperparameters}
  \begin{tabular*}{\textwidth}{@{\extracolsep{\fill}} lll}
    \hline
    Target$^{1}$                    & Hyperparameter                         & Search range$^{2,~3}$ \\
    \hline
    FOS                             & bin width                              & $p_1 \in \{4, 16, 64, 256\}$ \\ [0.8ex]
    \multirow{2}{*}{GLDS}           & bin width                              & $p_2 \in \{4, 16, 64, 256\}$ \\
                                    & distance                               & $p_3 \in \{1, 2, 3, 4\}$ \\ [0.8ex]
    \multirow{2}{*}{GLCM}           & bin width                              & $p_4 \in \{4, 16, 64, 256\}$ \\ 
                                    & distance                               & $p_5 \in \{1, 2, 3, 4\}$ \\ [0.8ex]
    GLRLM                           & bin width                              & $p_6 \in \{4, 16, 64, 256\}$ \\ [0.8ex]
    ADF                             & infinitesimal angle                    & $p_7 \in \{1, 2, 3, 4\}$ \\ [0.8ex]
    RDF                             & infinitesimal radius                   & $p_8 \in \{1, 2, 3, 4\}$ \\ [0.8ex]
    \hdashline
    FS                              & number of FS-selected features         & $p_{c_1} \in \{2, 3, 4, \cdots, 38\}$ \\ [0.8ex]
    FS + DC                         & number of FS-selected features with DC & $p_{c_2} \in \{3, 4, 5, \cdots, 38\}$ \\ [0.8ex]
    SVM with LK                     & cost parameter                         & $p_{c_3} \in [0.0001, 10000]$ \\ [0.8ex]
    \multirow{2}{*}{SVM with RBF}   & cost parameter                         & $p_{c_4} \in [0.0001, 10000]$ \\
                                    & $\gamma$ parameter                     & $p_{c_5} \in [0.0001, 10000]$ \\ [0.8ex]
    \multirow{2}{*}{DT}             & criterion                              & $p_{c_6} \in$ \{``gini'', ``entropy'', ``log loss''\}\\
                                    & max depth                              & $p_{c_7} \in \{1, 2, 3, 4, 5\}$ \\ [0.8ex]
    \hline
  \end{tabular*}

  \footnotesize{
  \begin{tablenotes}
    \item[$1$] FOS stands for first-order statistics.
    GLDS stands for gray level difference statistics.
    GLCM stands for gray level co-occurrence matrix.
    GLRLM stands for gray level run length matrix.
    ADF stands for angular distribution function via fourier power spectrum.
    RDF stands for radial distribution function via fourier power spectrum.
    FS stands for feature selection. 
    DC stands for dimensional compression.
    SVM stands for support vector machine.
    LK stands for linear kernel.
    RBF stands for radial basis function kernel.
    DT stands for decision tree.
    \item[$2$] $\{\cdot\}$ represents each discrete value or category, and $[A,B]$ represents a continuous value of $A$ or more and $B$ or less.
    \item[$3$] The subscript $c$ of the parameter variable varies depending on the element of the selected model vector (Equation~\ref{eq:model}).
    For example, if the combination is FS and SVM with LK ($m_{1}=\mathrm{FS}$, $m_{2}=\mathrm{None}$, $m_{3}=\mathrm{SVM~with~LK}$), it is $c_1=9$ and $c_3=10$. If the combination is FS + DC and DT ($m_{1}=\mathrm{FS}$, $m_{2}=\mathrm{DC}$, $m_{3}=\mathrm{DT}$), it is $c_2=9$, $c_6=10$, $c_7=11$.
  \end{tablenotes}
  }
  \end{threeparttable}
\end{table*}
%%%%%%%%%%%%%%%%%%%%%%%%%%%%%%%%%%%%

\section{Experiment} \label{sec:Experiment}
\subsection{Objectives and overview}
This study aims to verify the effectiveness of texture features in the pathological diagnosis of cardiomyopathy 
and to clarify a model design suitable for diagnosis with small sample size data.
These aspects are investigated using a three-class classification problem that diagnoses three patterns: 
two types of cardiomyopathy and the normal state.
Texture features are extracted from histopathological images of the myocardium.
The effectiveness of the texture features is verified by visualizing them 
and investigating the significance of the differences in class distribution using statistical hypothesis testing.
In addition, the model design most suitable for diagnosing cardiomyopathy using small sample data is clarified 
by comparing the prediction performance of each model design.

The experimental procedure is shown in 
Algorithms~\ref{algo:ModelDesignEvaluation},~\ref{algo:SearchBestParameters}, and~\ref{algo:DimensionalityReduction}.
First, the collection and processing of the dataset used in this experiment are explained.
Next, additional details on some of the variables and functions used in the algorithms are provided. 
Finally, a step-by-step explanation of the specific experimental procedure is given.

%%%%%%%%%  表：モデル設計の疑似コード，その1  %%%%%%%%%%
\begin{figure}[!t]
	\begin{algorithm}[H]
    \caption{An algorithm for evaluating the predictive performance of a model design.}
    \label{algo:ModelDesignEvaluation}
    \begin{algorithmic}[1]
      \Require{
        \Statex $D$: Dataset
        \Comment{Sec.~\ref{subsec:data}}

        \Statex $\bm{m}$: Model vector, $K$: Number of cross-validation splits, $B$: Iterations in Bayesian Optimization
        \Comment{Sec.~\ref{subsec:explanation}}
      }
      \Ensure{
        \Statex ($s^\mathrm{mean}_\mathrm{train}$, $s^\mathrm{std}_\mathrm{train}$),
        ($s^\mathrm{mean}_\mathrm{valid}$, $s^\mathrm{std}_\mathrm{valid}$),
        ($s^\mathrm{mean}_\mathrm{test}$, $s^\mathrm{std}_\mathrm{test}$): 
        Mean and standard deviation of macro-average F1-scores for each dataset
        \Statex \Comment{Tab.~\ref{table:performance evaluation}}
      }

      \State Dataset type: $J$ $\leftarrow$ \{$\mathrm{train}$, $\mathrm{valid}$, $\mathrm{test}$\}
      \State Initialize score sets: $S_{j} \leftarrow \varnothing$, $j \in J$
      \State Make subsets: $D_1$, $D_2$, $\cdots$, $D_K$ $\leftarrow$ $\mathrm{CrossVal}(D, K)$
      \Statex \Comment{Sec.~\ref{subsec:Cross-validation and hyperparameter optimization} \&~\ref{subsec:explanation}}
      
      \For{ $k \leftarrow 1$ \textbf{to} $K$ } 
        \State Make training set: $D_{\mathrm{train}} \leftarrow D \setminus D_{k}$
        \Comment{Fig.~\ref{fig:Cross-validation}}
        
        \State Make test set: $D_{\mathrm{test}} \leftarrow D_{k}$
        \Comment{Fig.~\ref{fig:Cross-validation}}

        \State $s_{\mathrm{valid}}$, $\bm{p}^{*}$ $\leftarrow$ SearchBestParameters($D_{\mathrm{train}}$, $K$, $B$, $\bm{m}$) 
        \Statex \Comment{Alg.~\ref{algo:SearchBestParameters}}

        \State Extract features: $F_{\mathrm{train}}$, $F_{\mathrm{test}}$ 
        \Statex \hspace{1.5em} $\leftarrow$ TextureFeatures($D_{\mathrm{train}}$, $D_{\mathrm{test}}$, $\bm{p}^{*}$) 
        \Comment{Sec.~\ref{subsec:Texture feature extraction}}

        \State Reduce features: $F_{\mathrm{train}}$, $F_{\mathrm{test}}$ 
        \Statex \hspace{1.5em} $\leftarrow$ DimensionalityReduction($F_{\mathrm{train}}$, $F_{\mathrm{test}}$, $\bm{m}$, $\bm{p}^{*}$) 
        \Statex \Comment{Alg.~\ref{algo:DimensionalityReduction}}

        \State Get scores: $s_{\mathrm{train}}$, $s_{\mathrm{test}}$ 
        \Statex \hspace{1.5em} $\leftarrow$ ClassifierScore($F_{\mathrm{train}}$, $F_{\mathrm{test}}$, $\bm{m}$, $\bm{p}^{*}$) 
        \Comment{Sec.~\ref{subsec:Classification models}}

        \State Merge scores: $S_j \leftarrow S_j \cup \{  s_j \}$, $j \in J$
      \EndFor
      \State Get means of scores: $s_{j}^\mathrm{mean} \leftarrow \mathrm{Mean}(S_j)$, $j \in J$
      \State Get standard deviations of scores: $s_{j}^\mathrm{std} \leftarrow \mathrm{Std}(S_j)$, $j \in J$
      \State \Return \{ ($s_{j}^\mathrm{mean}$, $s_{j}^\mathrm{std}$) $\mid$ $j \in J$\}
    \end{algorithmic}
	\end{algorithm}
\end{figure}
%%%%%%%%%%%%%%%%%%%%%%%%%%%%%%%%%%%%

%%%%%%%%%  表：モデル設計の疑似コード，その2  %%%%%%%%%%
\begin{figure}[!ht]
  \begin{algorithm}[H]
    \caption{Bayesian Optimization-based hyperparameter search with cross-validation.}
    \label{algo:SearchBestParameters}
    \begin{algorithmic}[1]
      \Require{
        \Statex $D_{\mathrm{train}}$: Training set, $\bm{m}$: Model vector, $K$: Number of cross-validation splits, $B$: Iterations in Bayesian Optimization
        \Statex \Comment{Sec.~\ref{subsec:explanation}}
      }
      \Ensure{
        \Statex $s^\mathrm{max}_\mathrm{valid}$: Maximum score of validation data, $\bm{p}^*$: Optimized hyperparameter vector
      }

      \State Make subsets: $D_1$, $D_2$, $\cdots$, $D_K$ $\leftarrow$ CrossVal($D_{\mathrm{train}}$, $K$)
      \Statex \Comment{Sec.~\ref{subsec:Cross-validation and hyperparameter optimization} \&~\ref{subsec:explanation}}

      \For{ $b \leftarrow 1$ \textbf{to} $B$ }
        \State Initialize validation score sets: $S_{\mathrm{valid}} \leftarrow \varnothing$

        \State $\bm{p} \leftarrow$ Draw the hyperparameter vector 
        \Statex \hspace{1.5em} from Table~\ref{table:hyperparameters} by Bayes. Opt.
        \Comment{Sec.~\ref{subsec:Cross-validation and hyperparameter optimization} \&~\ref{subsec:explanation}}

        \For{ $k \leftarrow 1$ \textbf{to} $K$ }
          \State Make split training set: $D_{\mathrm{st}} \leftarrow D_{\mathrm{train}} \setminus D_k$
          \Comment{Fig.~\ref{fig:Cross-validation}}

          \State Make validation set: $D_{\mathrm{valid}} \leftarrow D_k$
          \Comment{Fig.~\ref{fig:Cross-validation}}

          \State Extract features: $F_{\mathrm{st}}$, $F_{\mathrm{valid}}$ 
          \Statex \hspace{3em} $\leftarrow$ TextureFeatures($D_{\mathrm{st}}$, $D_{\mathrm{valid}}$, $\bm{p}$) 
          \Comment{Sec.~\ref{subsec:Texture feature extraction}}
          
          \State Reduce features: $F_{\mathrm{st}}$, $F_{\mathrm{valid}}$ 
          \Statex \hspace{3em} $\leftarrow$ DimensionalityReduction($F_{\mathrm{st}}$, $F_{\mathrm{valid}}$, $\bm{m}$, $\bm{p}$) 
          \Statex \Comment{Alg.~\ref{algo:DimensionalityReduction}}
          
          \State Get scores: $s_{\mathrm{valid}}$ 
          \Statex \hspace{3em} $\leftarrow$ ClassifierScore($F_{\mathrm{st}}$, $F_{\mathrm{valid}}$, $\bm{m}$, $\bm{p}$) 
          \Comment{Sec.~\ref{subsec:Classification models}}

          \State Merge scores: $S_{\mathrm{valid}} \leftarrow S_{\mathrm{valid}} \cup \{  s_{\mathrm{valid}} \}$
          \Comment{Sec.~\ref{subsec:Classification models}}
        \EndFor

        \State Record mean score: $s^{\mathrm{mean}}_{\mathrm{valid},\: b}$ $\leftarrow$ Mean($S_{\mathrm{valid}}$)

        \State Record hyperparameter: $\bm{p}_{b} \leftarrow \bm{p}$
      \EndFor

      \State Get optimal step id.: $b^*$ 
      \Statex $\leftarrow$ $\underset{b}{\mathrm{arg~max}}$ \{$s^{\mathrm{mean}}_{\mathrm{valid},\: b}$ $\mid$ $b$ $\in$ \{$1$, $2$, $\cdots$, $B$\}\}

      \State Get maximum validation score: $s^\mathrm{max}_\mathrm{valid} \leftarrow s^\mathrm{mean}_{\mathrm{valid},\: b^*}$
      
      \State Get optimal parameter: $\bm{p}^* \leftarrow \bm{p}_{b^*}$
      
      \State \Return $s^\mathrm{max}_\mathrm{valid}$, $\bm{p}^*$
    \end{algorithmic}
	\end{algorithm}
\end{figure}
%%%%%%%%%%%%%%%%%%%%%%%%%%%%%%%%%%%%

%%%%%%%%%  表：モデル設計の疑似コード，その3  %%%%%%%%%%
\begin{figure}[!ht]
  \begin{algorithm}[H]
    \caption{Feature selection and/or dimensional compression, or no processing, for texture features.}
    \label{algo:DimensionalityReduction}
    \begin{algorithmic}[1]
      \Require{
        \Statex $F_{a}$, $F_{b}$: Texture features of datasets $a$ and $b$}, $\bm{m}$: Model vector, $\bm{p}$: Hyperparameter vector
        \Comment{Sec.~\ref{subsec:explanation}
      }
      \Ensure{
        \Statex $F_{a}$, $F_{b}$: Texture features to which dimensionality reduction has not been applied or has been applied
      }

      \If{$m_1=\mathrm{FS}$} \Comment{Eq.~\ref{eq:m1}}
        \State Select features: $F_{a}$, $F_{b}$ 
        \Statex \hspace{1.5em} $\leftarrow$ FeatureSelection($F_{a}$, $F_{b}$, $\bm{m}$, $\bm{p}$) 
        \Comment{Sec.~\ref{subsec:Feature selection}}
      \EndIf
      \If{$m_2=\mathrm{DC}$} \Comment{Eq.~\ref{eq:m2}}
        \State Compress features: $F_{a}$, $F_{b}$ 
        \Statex \hspace{1.5em} $\leftarrow$ DimensionalCompression($F_{a}$, $F_{b}$) 
        \Comment{Sec.~\ref{subsec:Dimensional compression}}
      \EndIf
      \State \Return $F_{a}$, $F_{b}$
	  \end{algorithmic}
	\end{algorithm}
\end{figure}
%%%%%%%%%%%%%%%%%%%%%%%%%%%%%%%%%%%%

\subsection{Subjects and data description} \label{subsec:data}
Four patients histopathologically diagnosed with hypertrophic cardiomyopathy (HCM) at autopsy from January 1, 2020, to December 31, 2022, at Nihon University Itabashi Hospital and University Hospital were enrolled.
Among them, two patients with histopathologically confirmed cardiomyocyte disarray characteristics of HCM were considered 
disease cases, and two patients without cardiomyocyte disarray were considered borderline cases. 
In addition, two patients who died due to reasons other than cardiac disease and 
were autopsied during the abovementioned period were considered normal cases.
Histological sections 4 micrometers in thickness from myocardial specimens were prepared and stained with hematoxylin-eosin. 
Ten histological sections were imaged using a microscope equipped with a digital camera 
(Digital Microscopic System DP-70, Olympus Corporation Japan) for each of the six cases. 
In other words, there were six subjects (two disease cases, two borderline cases, and two normal cases), 
and the sample size was 60.
The protocol was approved by the Ethics Review Board of Nihon University Itabashi Hospital (RK-210914-18).

\begin{figure*}[!ht]
  \centering
  \begin{tabular}{lr}  % 2列にする
    \begin{subfigure}{0.47\textwidth}
        \centering
        \includegraphics[width=\linewidth]{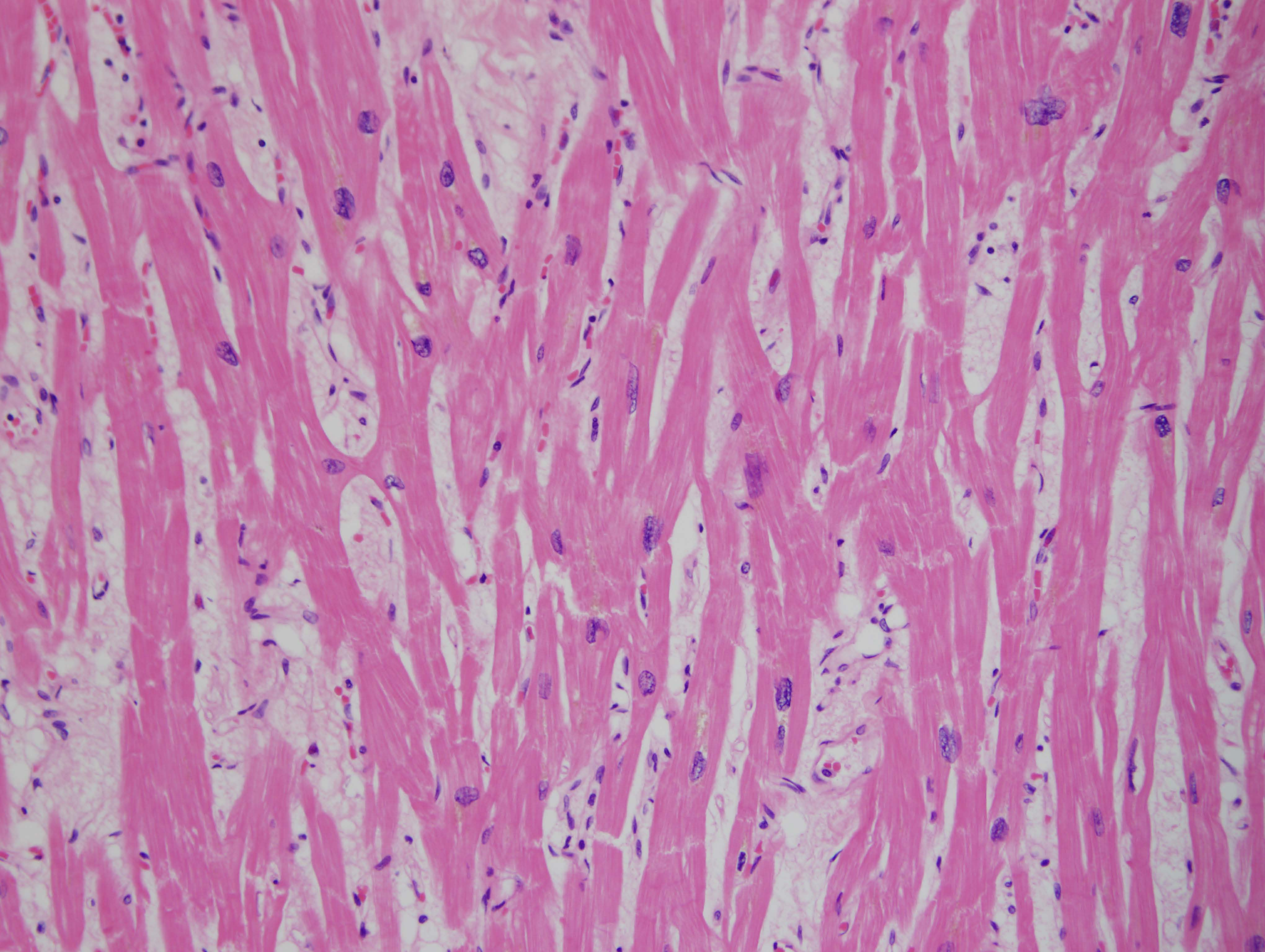}
        \caption{Disease case: Subject A}
    \end{subfigure} &
    \begin{subfigure}{0.47\textwidth}
        \centering
        \includegraphics[width=\linewidth]{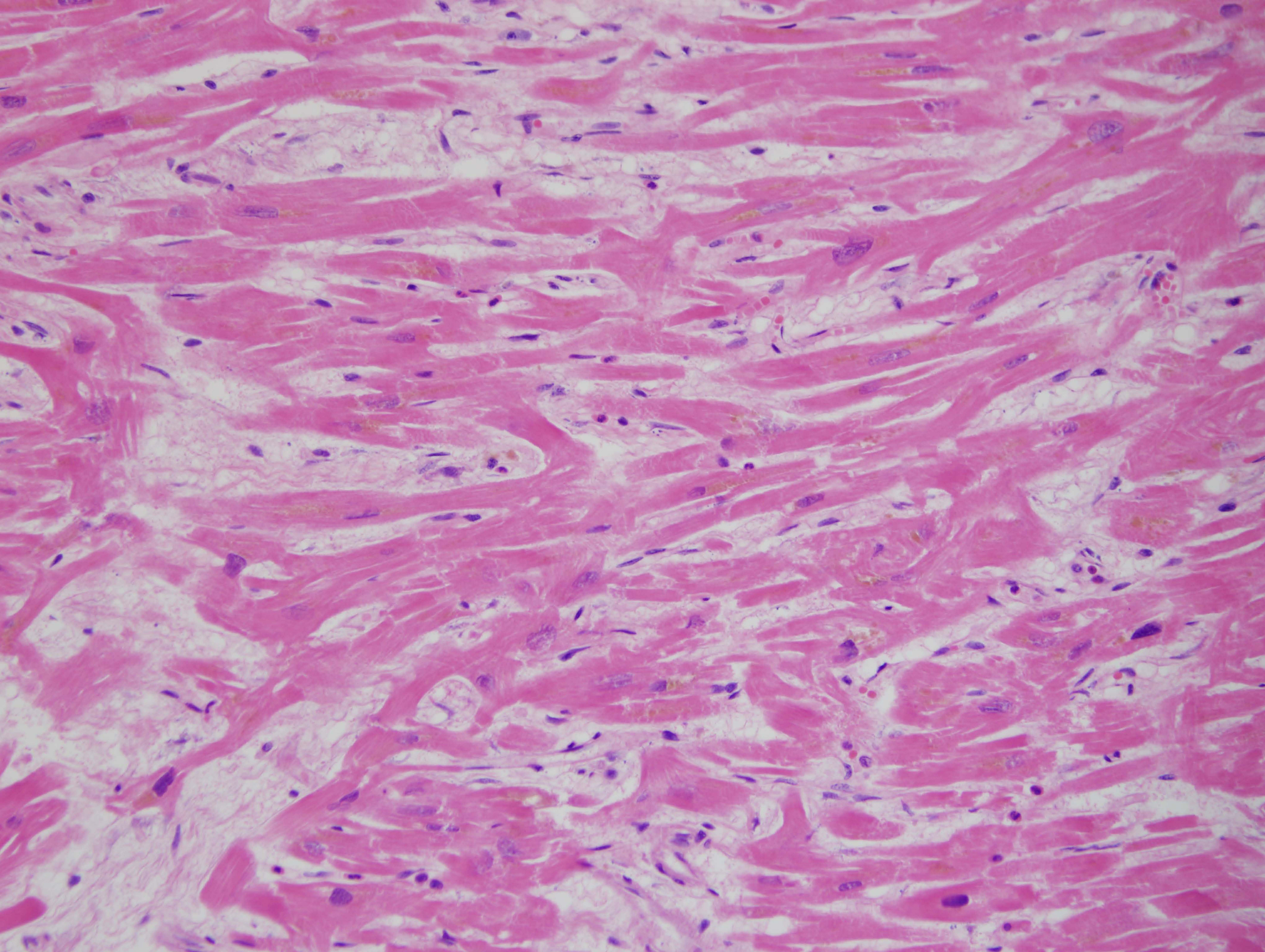}
        \caption{Disease case: Subject B}
    \end{subfigure} 
    \vspace{0.5cm} \\
    \begin{subfigure}{0.47\textwidth}
        \centering
        \includegraphics[width=\linewidth]{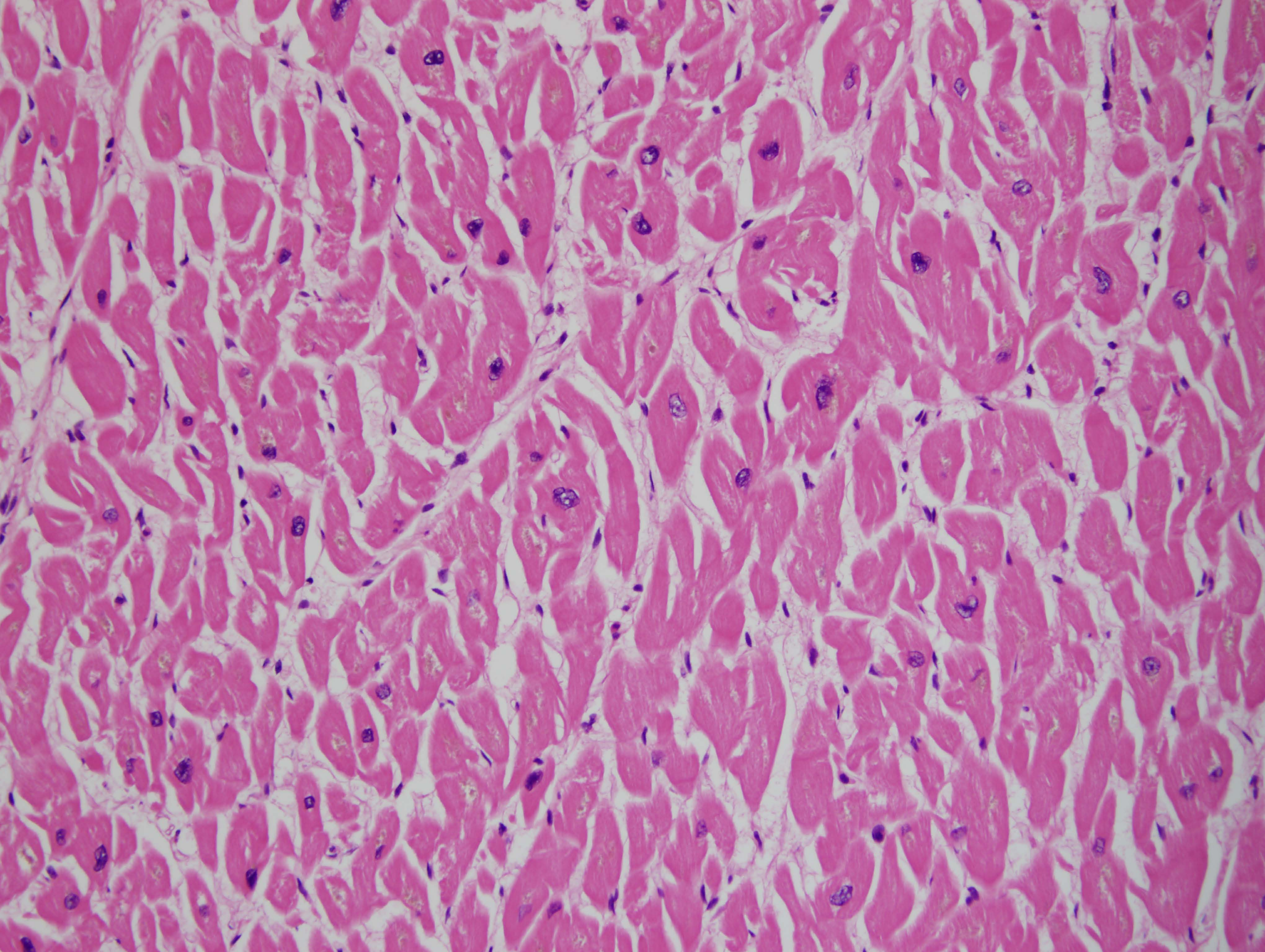}
        \caption{Borderline case: Subject C}
    \end{subfigure} &
    \begin{subfigure}{0.47\textwidth}
        \centering
        \includegraphics[width=\linewidth]{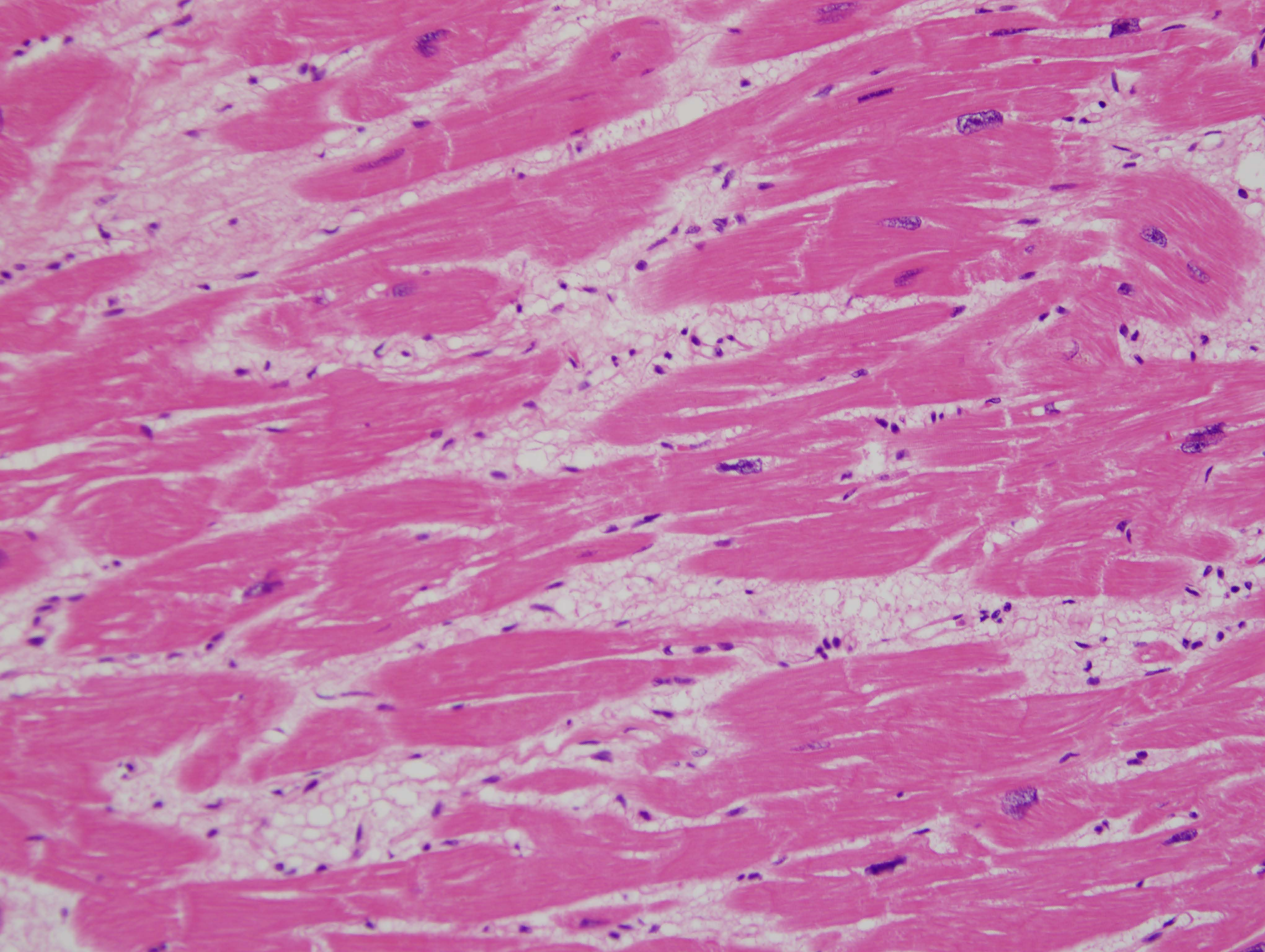}
        \caption{Borderline case: Subject D}
    \end{subfigure} 
    \vspace{0.5cm} \\
    \begin{subfigure}{0.47\textwidth}
        \centering
        \includegraphics[width=\linewidth]{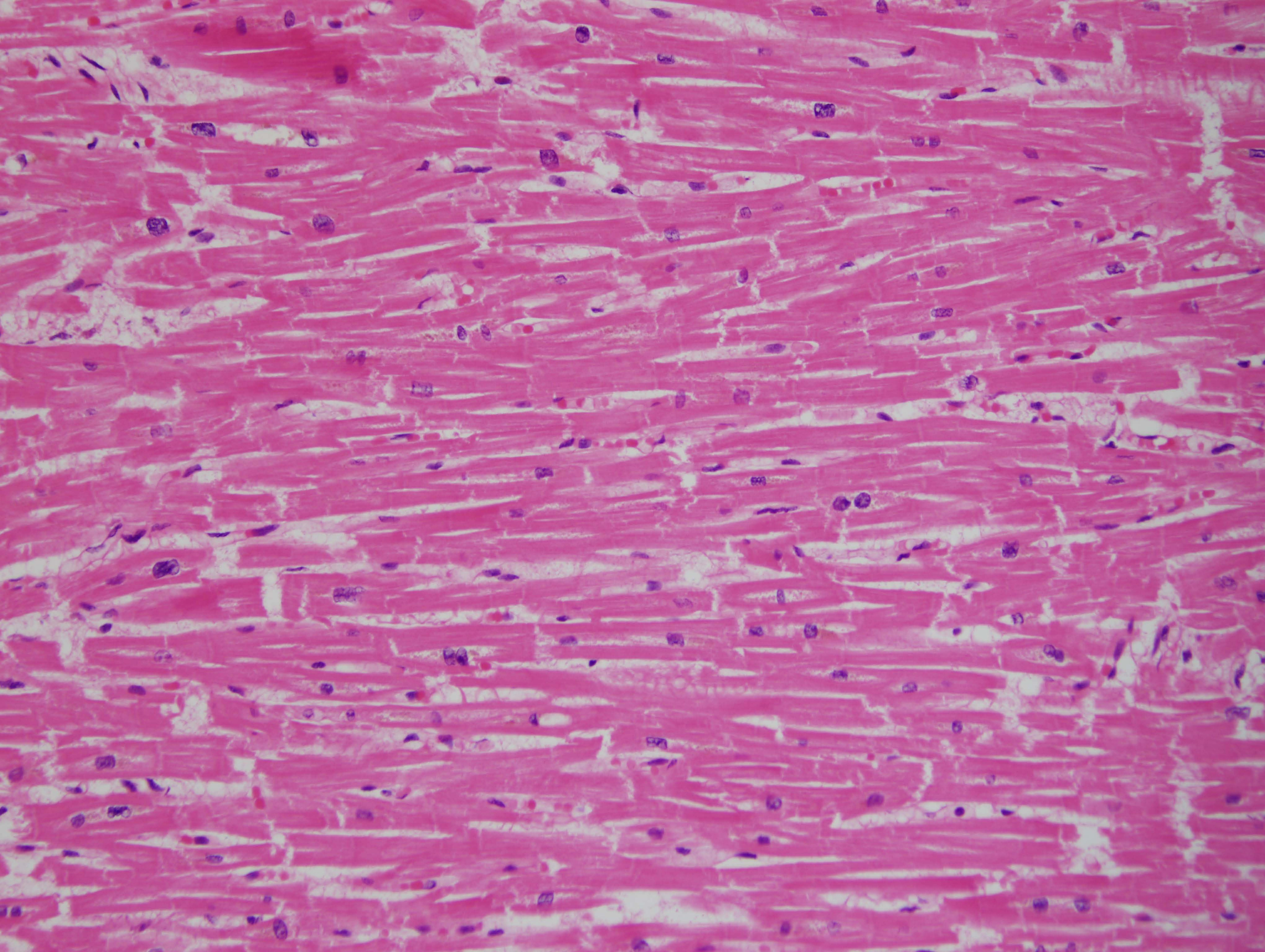}
        \caption{Normal case: Subject E}
    \end{subfigure} &
    \begin{subfigure}{0.47\textwidth}
        \centering
        \includegraphics[width=\linewidth]{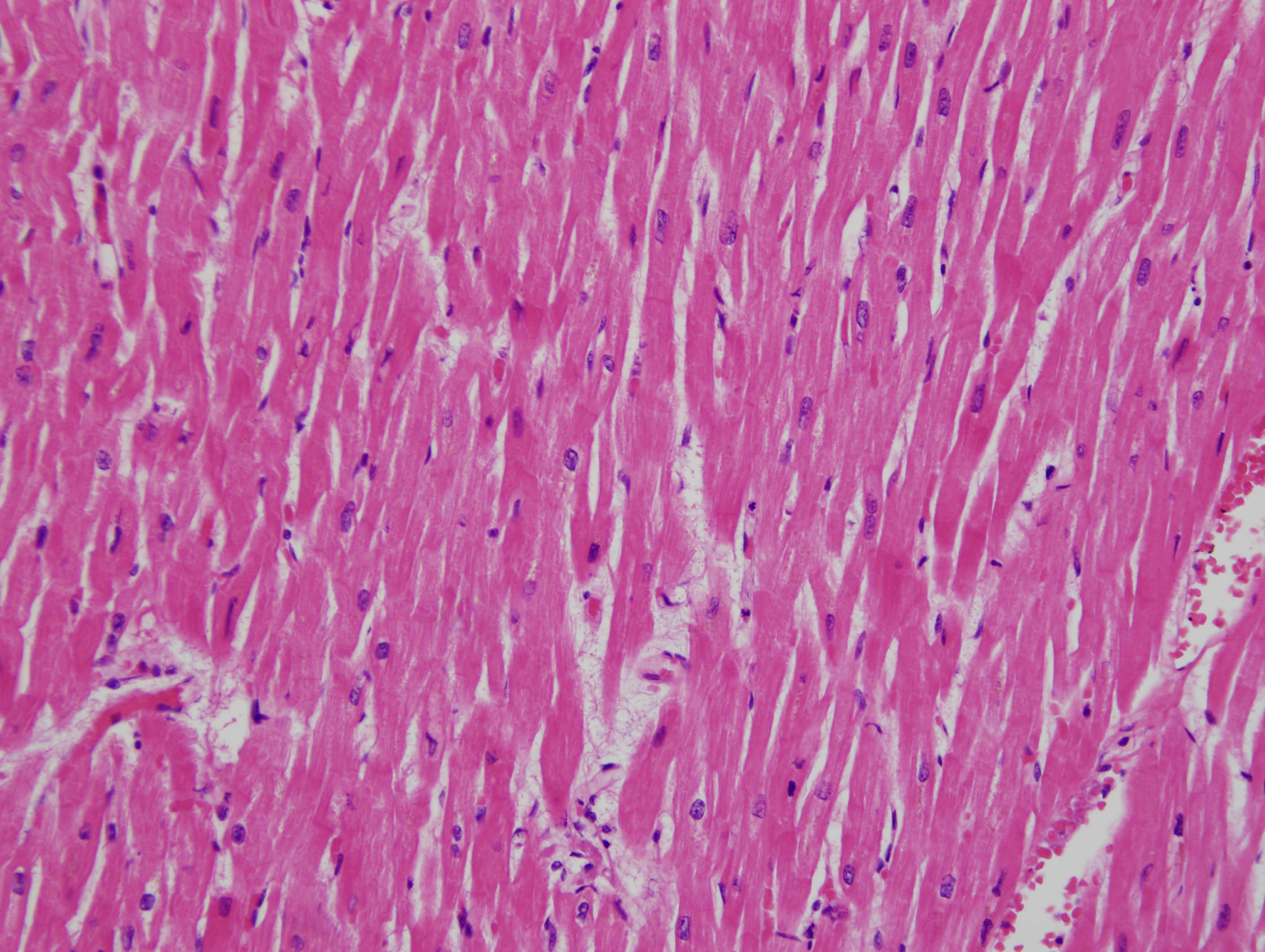}
        \caption{Normal case: Subject F}
    \end{subfigure}
  \end{tabular}
  \caption{Histopathological images of the myocardium from disease cases, borderline cases, and normal cases in six subjects.}
  \label{fig:histopathological image of the myocardium}
\end{figure*}

The histopathological images of the myocardium are shown in Figure~\ref{fig:histopathological image of the myocardium}.
These were 256-level color images composed of $4096\times3086\times3$ pixels each.
Because the similarity of the surrounding pixels is high, the calculation cost increases unnecessarily.
Therefore, for efficient image analysis, these images were converted to $204\times154\times1$ pixel size grayscale images.

The dataset $D$ used in this experiment is a set of vectors 
where the elements are pairs of converted grayscale images and their corresponding myocardial states (class labels).

%%%%%%%%%  図：心筋病理画像  %%%%%%%%%%
%%%%%%%%%%%%%%%%%%%%%%%%%%%%%%%%%%%%

\subsection{Explanation of variables and functions in Algorithms} \label{subsec:explanation}
First, the model vector $\bm{m}$ that represents the design of each model is defined.
In this study, the model vector is defined as,
\begin{align}
  \bm{m} = \begin{bmatrix} m_1 & m_2 & m_3 \end{bmatrix}^\top \in M. \label{eq:model}
\end{align}
However, 
\begin{align}
  m_{1} &\in M_1 := \{\mathrm{FS}, \mathrm{AF}\}, \label{eq:m1} \\
  m_{2} &\in M_2 := \{\mathrm{DC}, \mathrm{None}\}, \label{eq:m2} \\
  m_{3} &\in M_3 := \{\mathrm{SVM~with~LK}, \mathrm{SVM~with~RBF}, \mathrm{DT} \}, \\
  M &= M_1 \times M_2 \times M_3.
\end{align}
For example, $\bm{m} = \begin{bmatrix} \mathrm{FS} & \mathrm{None} & \mathrm{SVM~with~RBF} \end{bmatrix}^\top$ means that 
FS was performed, no DC was applied, and SVM using RBF was used.
In addition, $\bm{m} = \begin{bmatrix} \mathrm{AF} & \mathrm{DC} & \mathrm{DT} \end{bmatrix}^\top$ means that 
DT was used for features to which DC was applied across all features.
The number of patterns that $\bm{m}$ can be used is $|M| = |M_1| \times |M_2| \times |M_3| = 2\times 2\times 3 = 12$.
Therefore, there are 12 model vectors $\bm{m}_1$, $\bm{m}_2$, $\cdots$, $\bm{m}_{12}$ $\in$ $M$.

Next, we provide additional details on the CrossVal function, 
which is a dataset partitioning function based on stratified $K$-fold cross-validation.
Let $D$ be a dataset comprising pairs of input images and class labels. 
This dataset $D$ is divided into $K$ subsets of equal size using the function
\begin{align}
  D_1, D_2, \cdots, D_K = \mathrm{CrossVal}(D, K). \label{eq:CrossVal}
\end{align}
Here, 
\begin{align}
  |D_1| = |D_2| = \cdots = |D_K|, \ D = \bigcup_{k=1}^{K} D_k,
\end{align}
is assumed.
That is, all subsets $D_k$ have the same number of data samples, and there is no data overlap among the different subsets. 
In this case, the class-label distribution remained uniform across all subsets 
(Sections~\ref{subsec:Cross-validation and hyperparameter optimization}).
The number of splits, denoted as $K$, used for the stratified $K$-fold cross-validation represents the number of 
data blocks used to divide the dataset $D$ into training, validation, and test sets (Figure~\ref{fig:Cross-validation}).
In this experiment, the value of $K$ was set to five 
for both the training--test and the split training--validation splits, as shown in Figure~\ref{fig:Cross-validation}.
Therefore, the training set $D_{\mathrm{train}}$ and test set $D_{\mathrm{test}}$ were divided 
at a ratio of 8:2 (Algorithm~\ref{algo:ModelDesignEvaluation}: Lines 5--6).
In addition, the split training set $D_{\mathrm{st}}$ and validation set $D_{\mathrm{valid}}$ were divided 
at a ratio of 8:2 (Algorithm~\ref{algo:SearchBestParameters}: Lines 6--7).
In other words, the training and test sets were evaluated five times, and the validation set was evaluated 25 times.

Finally, the hyperparameter vector $\bm{p}$, which corresponds to the texture features, 
FS, and the classification model in Table~\ref{table:hyperparameters}, is defined.
It is represented by 
\begin{align}
  \bm{p} = \begin{bmatrix} p_1 & p_2 & p_3 & \cdots \end{bmatrix}^\top. \label{eq:hypara}
\end{align}
In Table~\ref{table:hyperparameters}, the hyperparameters of the texture features, 
bin width for FOS, GLDS, GLCM, and GLRLM ($p_1$, $p_2$, $p_4$, $p_6$), were set to four values (4, 16, 64, and 256), 
which were powers of four in each case.
In addition, for the distance of GLDS and GLCM ($p_3$, $p_5$), four values from one to four were adopted for each.
Furthermore, for infinitesimal angle/radius of ADF and RDF ($p_7$, $p_8$), four values from one to four were used for each.
For the number of FS-selected features ($p_{c_1}$), a hyperparameter for FS, 37 values from 2 to 38 were used.
In this study, 39 features are extracted regardless of the image or sample size of the histopathological images of the myocardium (Table~\ref{table:texture features}).
Therefore, the number of dimensions to be selected ranged from 2 to 38, 
which is the minimum number of dimensions in which the interaction of the features can be considered.
However, when DC is applied, the number of dimensions is two, as described in Section~\ref{subsec:Dimensional compression}.
Thus, in the number of FS-selected features with DC ($p_{c_2}$), the number of dimensions for FS is set to three or more.
The cost parameter ($p_{c_3}$, $p_{c_4}$) and $\gamma$ parameter ($p_{c_5}$), 
which are hyperparameters of the classification model, 
were adopted in the range of $1\times10^{-4}$ to $1\times10^{4}$.
For criterion ($p_{c_6}$), three values of gini, entropy, and log loss were used, 
and for max depth ($p_{c_7}$), five values from one to five were used.

The numbers of $p_1$, $p_2$, $\cdots$, $p_8$ variables, which are hyperparameters of the texture features, are fixed and independent of the model design.
In contrast, the number of $p_{c_1}$, $p_{c_2}$, $\cdots$, $p_{c_7}$ hyperparameters for the FS and classification models 
depends on the model vector $\bm{m}$ and changes dynamically according to the elements of the selected model vector.
For example, if $\bm{m} = \begin{bmatrix} \mathrm{AF} & \mathrm{None} & \mathrm{SVM~with~LK} \end{bmatrix}^\top$, 
then $p_{c_3}$ is adopted as the hyperparameter of the classification model, resulting in $c_3=9$.
That is, the hyperparameter $\bm{p}$ is nine-dimensional.
For $\bm{m} = \begin{bmatrix} \mathrm{FS} & \mathrm{DC} & \mathrm{DT} \end{bmatrix}^\top$, 
$p_{c_2}$ is used as the hyperparameter for feature selection, and $p_{c_6}$ and $p_{c_7}$ are used 
as the hyperparameters of the classification model, resulting in $c_2=9$, $c_6=10$, $c_7=11$.
In other words, the hyperparameter $\bm{p}$ is 11-dimensional.

The hyperparameter vector thus defined is then optimized using Bayesian optimization.
In this experiment, the number of iterations in the Bayesian optimization $B$ was set as 300.

\subsection{Experimental procedure with Algorithms}
In this experiment, the mean and standard deviation 
$s^\mathrm{mean}_j$, $s^\mathrm{std}_j$, $j$ $\in$ \{ $\mathrm{train}$, $\mathrm{valid}$, $\mathrm{test}$ \} 
of the macro-average F1-score for each dataset were calculated using dataset $D$, model vector $\bm{m}$, 
number of cross-validation splits $K$, and iterations in Bayesian optimization $B$ as the inputs.
First, set $J$ is defined to represent each dataset, 
and an empty set $S_{j}$, $j \in J$ is defined to store the output evaluation values. 
Dataset $D$ is partitioned into subsets $D_1$, $D_2$, $\cdots$, $D_K$ based on the 
stratified $K$-fold cross-validation (Equation~\ref{eq:CrossVal}).
One block of $K$ partitions is then assigned to the test set $D_{\mathrm{test}}$ 
and the remaining $K-1$ blocks are assigned to the training set $D_{\mathrm{train}}$.
Then, the SearchBestParameters function is executed to search for hyperparameters suitable for the dataset and model design 
(Algorithm~\ref{algo:ModelDesignEvaluation}: Lines 1--7).

In the SearchBestParameters function in Algorithm~\ref{algo:SearchBestParameters}, 
given input $D_{\mathrm{train}}$, $K$, $B$, and $\bm{m}$, 
are returned the score $S^{\mathrm{max}}_{\mathrm{valid}}$ and the hyperparameter $\bm{p}^{*}$ 
when the average generalization performance on the validation set was the highest, 
using a combined cross-validation and hyperparameter optimization method.
Here, the training set $D_\mathrm{train}$ is first partitioned into the subsets $D_1$, $D_2$, $\cdots$, $D_K$ 
based on stratified $K$-fold cross-validation (Equation~\ref{eq:CrossVal}).
Subsequently, an empty set $S_\mathrm{valid}$ is declared to store the validation score obtained in one optimization run.
Furthermore, a hyperparameter vector $\bm{p}$ is constructed from the search range listed in Table~\ref{table:hyperparameters} 
using Bayesian optimization according to Equation~\ref{eq:hypara} (Algorithm~\ref{algo:SearchBestParameters}: Lines 1--4).
Next, in the $K$ subsets, one block is assigned to the validation set $D_{\mathrm{valid}}$ 
and the remaining $K-1$ blocks were assigned to the split training set $D_{\mathrm{st}}$.
Texture feature extraction is then performed on $D_{\mathrm{st}}$ and $D_{\mathrm{valid}}$ 
based on hyperparameter vector $\bm{p}$.
In the texture feature extraction, 39 features were first extracted, and then robust standardization was applied.
The median and interquartile range were calculated using the $F_{\mathrm{st}}$ 
and robust standardization was applied to $F_{\mathrm{st}}$ and $F_{\mathrm{valid}}$ based on these values 
(Algorithm~\ref{algo:SearchBestParameters}: Lines 6--8).

For dimensionality reduction, FS, DC, and both are applied or not, depending on the model design 
(Algorithm~\ref{algo:SearchBestParameters}: Line 9).
Therefore, in the DimensionalityReduction function in Algorithm~\ref{algo:DimensionalityReduction}, 
the dimensionality reduction method was selected based on whether the model vector $\bm{m}$ contains FS and/or DC parameters.
If DC is applied, the transformation parameters are determined according to the sample distribution of the input data.
Thus, the $F_{\mathrm{st}}$ is compressed first, followed by the $F_{\mathrm{valid}}$ 
using the transformation parameters calculated from $F_{\mathrm{st}}$.
If the model vector $\bm{m}$ does not contain FS and DC, 
dimensionality reduction is not applied and the feature inputs to the DimensionalityReduction function are returned 
as they are (Algorithm~\ref{algo:DimensionalityReduction}: Lines 1--7).

Based on the feature vectors $F_{\mathrm{st}}$ and $F_{\mathrm{valid}}$ returned by the DimensionalityReduction function, 
$\bm{m}$ indicating the classification model to be applied, and its hyperparameter $\bm{p}$, 
the evaluation value $s_{\mathrm{valid}}$ for the validation set is calculated.
Subsequently, $K$ instances of $s_{\mathrm{valid}}$ are merged into $S_{\mathrm{valid}}$.
The mean values $s^{\mathrm{mean}}_{\mathrm{valid},\: b}$ and their hyperparameter $\bm{p}_{b}$ are stored 
as the numerical value at the $b$th Bayesian optimization iteration (Algorithm~\ref{algo:SearchBestParameters}: Lines 10--14).
When the procedure up to this point is performed $B$ times, hyperparameter $b^*$ is extracted 
when $s^{\mathrm{mean}}_{\mathrm{valid},\: b}$ is the largest.
A validation score $s^{\mathrm{mean}}_{\mathrm{valid},\: b^*}$ is adopted 
as the maximum score $S^{\mathrm{max}}_{\mathrm{valid}}$.
Furthermore, the hyperparameter $\bm{p}_{b^*}$ at the $b^*$-th iteration is adopted 
as the optimized hyperparameter $\bm{p}^*$ (Algorithm~\ref{algo:SearchBestParameters}: Lines 16--19).

Extraction of texture features and dimensionality reduction and calculation of the classification scores 
for the training set $D_{\mathrm{train}}$ and test set $D_{\mathrm{test}}$ in Algorithm~\ref{algo:ModelDesignEvaluation} 
were performed using the optimized hyperparameters $\bm{p}^*$.
This procedure follows lines 8--11 of Algorithm~\ref{algo:SearchBestParameters}.
Consequently, sets of evaluation metrics, $S_{\mathrm{train}}$, $S_{\mathrm{valid}}$, and $S_{\mathrm{test}}$ 
were obtained for the training, validation, and test sets, respectively.
Subsequently, the mean and standard deviation of these metrics were returned 
as the prediction performance for a given model design $\bm{m}$ (Algorithm~\ref{algo:ModelDesignEvaluation}: Lines 8--15).

%%%%%%%%%  図：箱ひげ図  %%%%%%%%%%
\begin{figure*}[!ht]
  \centering
  \includegraphics[width=\textwidth]{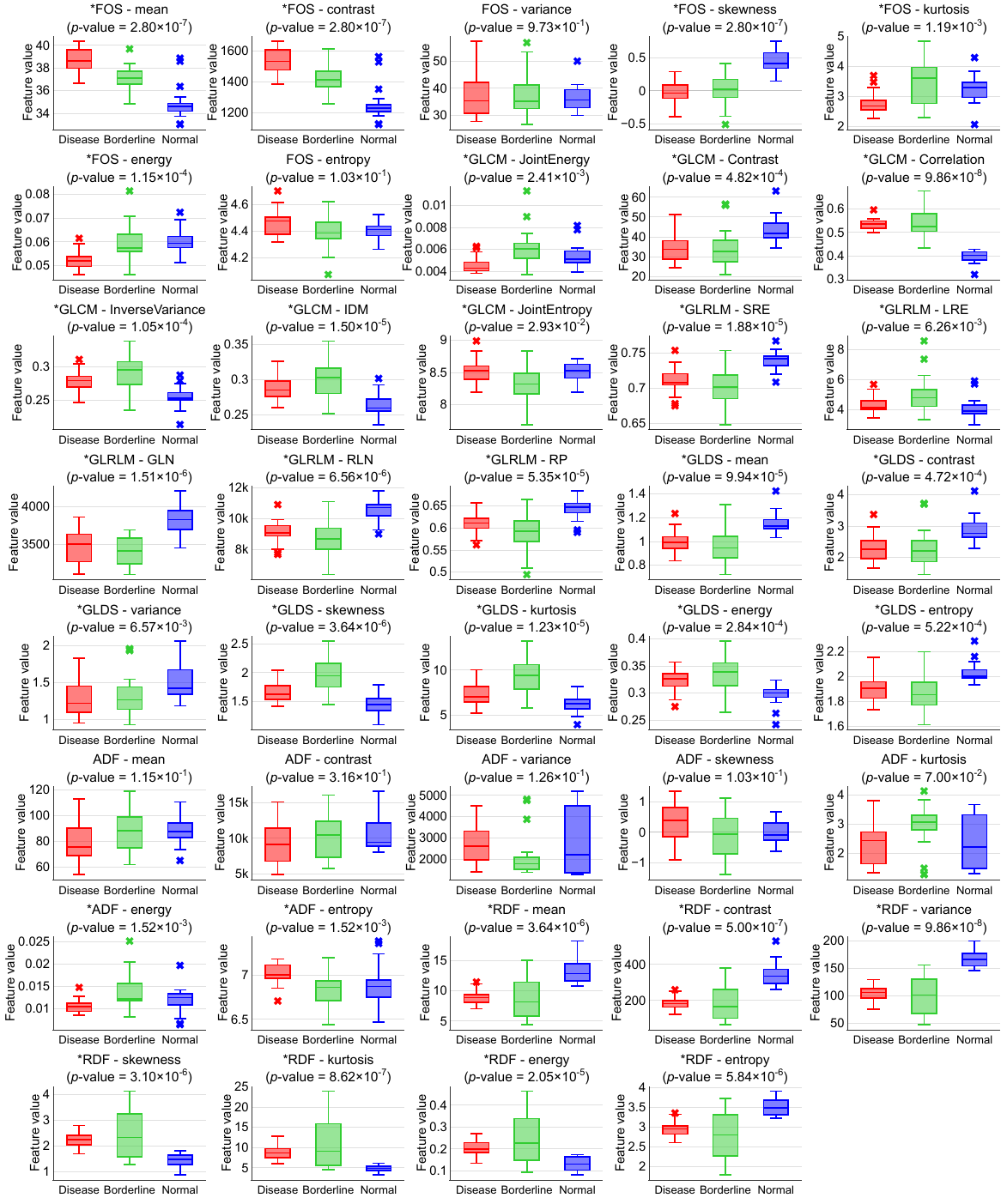}
  \caption{Box plots for disease, borderline, and normal cases across all texture features, and adjusted $p$-values for the three groups from statistical hypothesis testing (Features with significant differences are marked with an asterisk beside the variable name).}
  \label{fig:boxplot}
\end{figure*}
%%%%%%%%%%%%%%%%%%%%%%%%%%%%%%%%%%%%

\section{Results and discussion} \label{sec:Results and discussion}
\subsection{Effectiveness of texture features for myocardial pathology diagnosis}\label{subsec: result of texture features}
First, the effectiveness of texture features in diagnosing myocardial pathology was evaluated for the disease, borderline, and normal case categories.
Thus, the set of optimal hyperparameters $\bm{p}^{*}$ ($|M| \times K = 60$), 
employed in each cross-validation for all model designs, $M$, was collected.
The most frequently adopted values for each element were aggregated, 
and texture features were extracted using the reconstructed hyperparameter vector.
Statistical hypothesis testing was conducted to examine whether texture features exhibited distinct class distributions in one or more diseased, borderline, or normal cases.
The statistical hypothesis testing methods employed were 
the Kruskal-Wallis test, a nonparametric test designed for three or more groups, 
and the Benjamini-Hochberg method, a multiple-testing correction.
$\alpha=0.05$ was set as the significance level.
Box plots of these texture features and their adjusted $p$-values are shown in Figure~\ref{fig:boxplot}.

Here, it was qualitatively confirmed that while a few features clearly distinguished all three classes, 
many features exhibited a clear separation between one class and the others.
Among these features, the mean of FOS indicated that the three classes were clearly separated.
This metric represents the centroid of the gray level distribution in an image, and distinct trends were observed across the classes.
Specifically, in disease cases where the median of the box plot was the highest, 
the image shows a stronger prevalence of whites compared to the other classes.
The borderline case showed a slightly lower median, whereas the normal case, with the lowest median, 
demonstrated a stronger presence of black across the entire image.
This trend is also confirmed in Figure~\ref{fig:histopathological image of the myocardium}.

More than half of the features exhibited statistically significant differences, 
with significance confirmed for 32 of the 39 ($\approx 82.1\%$) features, 
excluding the variance and entropy of FOS and the mean, contrast, variance, skewness, and kurtosis of ADF.
This suggests that texture features have the potential to extract distinct trends in the three myocardial conditions.
However, for many statistical metrics of ADF, no significant differences were observed in the class distributions.
Therefore, it is suggested that considering periodicity with respect to angular directionality may not be effective for the diagnosis of myocardial pathology.
This may be due to a lack of uniformity in the fiber direction of the myocardial cells during image acquisition 
(Figure~\ref{fig:histopathological image of the myocardium}).
Thus, although the cost of data preparation is high, setting rules for the fiber orientation 
during image acquisition could potentially yield significant differences in the statistical metrics related to ADF.

%%%%%%%%%  表：モデル性能の比較  %%%%%%%%%%
\begin{table*}[t]
  \centering
  %{\small
  \begin{threeparttable}[h] % 注釈
  \caption{Macro-average F1-score (mean ± 1 standard deviation) for each dataset and classification model obtained using the dimensionality reduction algorithm, and average feature dimension after dimensionality reduction.}
  \label{table:performance evaluation}

  \begin{tabular*}{\textwidth}{@{\extracolsep{\fill}} lclccc}
    \hline
    \multicolumn{1}{c}{Classification} & Dimensionality &    \multicolumn{1}{c}{Feature}    & \multicolumn{3}{c}{Macro-average F1-score$^4$} \\
    \multicolumn{1}{c}{models$^1$}& Reduction$^2$  & \multicolumn{1}{c}{dimension$^3$} & Training set & Validation set & Test set \\
    \hline
    \multirow{4}{*}{SVM with LK}  &AF     & $39.0$                 & $.992 \pm .017$ & $.887 \pm .117$ & $.916 \pm .053$ \\
                                  &FS     & $39.0 \overset{\text{FS}}{\rightarrow} 32.4$        & $.992 \pm .017$ & $.880 \pm .113$ & $.867 \pm .040$ \\
                                  &DC     & $39.0 \overset{\text{DC}}{\rightarrow} 2.0$         & \underline{$\mathbf{1.00 \pm .000}$} & $.789 \pm .135$ & $.796 \pm .036$ \\
                                  &FS + DC& $39.0 \overset{\text{FS}}{\rightarrow} 18.4 \overset{\text{DC}}{\rightarrow} 2.0$& $.996 \pm .008$ & \underline{$.895 \pm .104$} & \underline{$\mathbf{.949 \pm .067}$}\\
    \hdashline
    \multirow{4}{*}{SVM with RBF} &AF     & $39.0$                 & $.992 \pm .017$ & \underline{$\mathbf{.906 \pm .076}$} & $.898 \pm .084$ \\
                                  &FS     & $39.0 \overset{\text{FS}}{\rightarrow} 34.0$        & $.996 \pm .008$ & $.900 \pm .131$ & $.882 \pm .041$ \\
                                  &DC     & $39.0 \overset{\text{DC}}{\rightarrow} 2.0$         & \underline{$\mathbf{1.00 \pm .000}$} & $.750 \pm .186$ & $.768 \pm .108$ \\
                                  &FS + DC& $39.0 \overset{\text{FS}}{\rightarrow} 18.2 \overset{\text{DC}}{\rightarrow} 2.0$& $.992 \pm .010$ & $.882 \pm .089$ & \underline{$.932 \pm .034$} \\
    \hdashline
    \multirow{4}{*}{Decision tree}&AF     & $39.0$                 & $.996 \pm .008$ & $.806 \pm .123$ & $.774 \pm .088$ \\
                                  &FS     & $39.0 \overset{\text{FS}}{\rightarrow} 36.2$        & $.983 \pm .024$ & \underline{$.811 \pm .141$} & $.754 \pm .090$ \\
                                  &DC     & $39.0 \overset{\text{DC}}{\rightarrow} 2.0$         & \underline{$\mathbf{1.00 \pm .000}$} & $.722 \pm .183$ & $.742 \pm .106$ \\
                                  &FS + DC& $39.0 \overset{\text{FS}}{\rightarrow} 18.2 \overset{\text{DC}}{\rightarrow} 2.0$& $.996 \pm .008$ & $.782 \pm .119$ & \underline{$.811 \pm .108$} \\
    \hline
  \end{tabular*}

  \footnotesize{
  \begin{tablenotes}
    \item[$1$] SVM stands for support vector machine. LK stands for linear kernel. RBF stands for radius basis function kernel.
    \item[$2$] AF stands for all features. FS stands for feature selection. DC stands for dimensional compression.
    \item[$3$] This presents the average number of dimensions after dimensionality reduction across five-fold cross-validation. The leftmost column lists the number of dimensions at input, while the columns to the right display the dimensions after dimensionality reduction (FS or/and DC).
    \item[$4$] \underline{The underlined values} indicate the highest classification performance within each classification model for each data set. \textbf{The bold values} indicate that the classification performance is the highest among all classification models for each data set.
  \end{tablenotes}
  }
  \end{threeparttable}
  %}
\end{table*}
%%%%%%%%%%%%%%%%%%%%%%%%%%%%%%%%%%%%

\subsection{Performance comparison of different model designs}
Next, the prediction performance of each model design $M$ is compared to identify the model design 
that is most suitable for the pathological diagnosis of cardiomyopathy using a small sample size.
The number of spatial dimensions after dimensionality reduction and the macro-average F1-score 
for each dataset for each model design are summarized in Table~\ref{table:performance evaluation}.

Initially, we focused on overall evaluation values: 
the minimum was 0.983 and the maximum was 1.000 for the training set, 
the minimum was 0.722 and the maximum was 0.906 for the validation set, 
and the minimum was 0.742 and the maximum was 0.949 for the test set.
From these results, it was confirmed that in all model designs, good overall prediction performance was achieved.
However, when focusing on the DC method, it was observed that in all the models, 
the fitting performance of the training set was 1.000, 
while the generalization performance on the validation set ranged from 0.722 to 0.789, 
and the generalization performance on the test set ranged from 0.742 to 0.796, indicating a tendency toward overfitting.
This suggests that the LDA, being a supervised dimensionality-reduction method, 
may have resulted in overfitting during DC.
This was not attributed to features that could be noise and did not contribute to the classification, 
as indicated by the results of the statistical hypothesis tests, 
but a rather high spatial dimensionality relative to the sample size.
A known limitation of classical LDA is that, as the number of features increases relative to the sample size, 
It becomes difficult to reliably estimate the within-class covariance, 
which has been reported as a cause of overfitting~\cite{qiao2008effective}.
In this experiment, cross-validation was applied to the dataset; therefore, the sample size of the split training set was 38.
In contrast, the number of features was 39, resulting in high-dimensional, low-sample size data 
(sample size $\ll$ features) ($38/39 \approx 0.97$).
Therefore, it is possible that the distance calculation between samples did not function properly.
To address this issue, combining unsupervised dimensionality reduction methods, 
such as principal component analysis~\cite{yang2021improving} or applying a regularized LDA, 
which has been extended to high-dimensional low-sample size data~\cite{friedman1989regularized}, is considered necessary.

Next, focusing on evaluation values other than the DC, 
SVM consistently exhibited better generalization performance than DT across all dimensionality reduction methods.
Within each model, a comparison of the generalization performance on the validation set revealed the following: 
AF, FS, and FS + DC exhibited similar performances based on their standard deviations.
However, on the test set, FS + DC achieved the highest generalization performance among all models, 
followed by AF in second place and FS in third place.

Generally, when the number of features is large relative to the sample size, 
overfitting is more likely to occur, which often results in a lower generalization performance~\cite{mori2022prediction}.
Therefore, in this experiment, AF was expected to exhibit a relatively lower generalization performance.
However, as illustrated in Figure~\ref{fig:boxplot} and described in Section~\ref{subsec: result of texture features}, 
32 of the 39 ($\approx 82.1\%$) features were considered useful for classification, 
indicating that the classes were likely separated in a high-dimensional space.
In addition, this study employed a nested structure for cross-validation and hyperparameter optimization, 
where the training and test sets as well as the split training and validation sets were hierarchically separated.
Nested cross-validation has been reported to be more robust against overfitting in small sample size datasets 
compared with conventional cross-validation~\cite{vabalas2019machine}.
These findings indicate that the extraction of highly predictive features, 
along with the use of cross-validation and hyperparameter optimization methods, which are robust to small sample sizes, 
enabled AF to achieve high generalization performance while avoiding overfitting, 
despite the large number of features relative to the sample size.

Focusing on the feature dimensions in FS, the number of dimensions after FS was 
32.4 for SVM with LK, 34.0 for SVM with RBF, and 36.2 for the DT, 
indicating that only features that could act as noise without contributing to the classification were eliminated.
However, cases in which FS outperformed AF in terms of generalization performance were rarely observed.
Thus, it can be inferred that when most features are useful, the spatial dimensionality was not notably reduced through FS. 
The removal of features that can act as noise without contributing to the classification has minimal impact on improving the generalization performance.

However, focusing on the feature dimensions of FS+DC, 
it was confirmed that after FS, the number of dimensions was approximately 18 for all models.
As described in Section~\ref{subsec: result of texture features} and shown in Figure~\ref{fig:boxplot}, 
it was recognized that 32 of the 39 features were useful for classification.
This indicates that not only features that may become noise and contribute little to classification, 
but also features that were important for classification, have been excluded.
In contrast, when focusing on the evaluation values of the test set, 
it was confirmed that FS+DC had the highest generalization performance across all the models.
The number of dimensions was approximately 18, which can be interpreted 
as the dimension where LDA can avoid overfitting and demonstrate its true discriminative power 
when the number of features was slightly larger than the sample size ($38/18 \approx 2.11$).
Furthermore, the fact that the number of dimensions after applying FS is similar across all models indicates that 
this is not model-dependent but rather specific to the dataset.
Thus, we demonstrated that the multi-step dimensionality-reduction process is effective 
when the number of features exceeded the sample size.
Moreover, this dimensionality-reduction method may be effective regardless of the decision boundary, 
particularly when the complexity of the model is low.

Table~\ref{table:selected texture features} lists the features selected through FS and their proportions 
based on five-fold cross-validation of the training and test sets 
for $\begin{bmatrix} \mathrm{FS} & \mathrm{DC} & \mathrm{SVM~with~LK} \end{bmatrix}^\top$, 
which exhibited the highest generalization performance for the test set.
By reviewing Table~\ref{table:selected texture features}, 
it was confirmed that many features exhibited significant differences 
in Figure~\ref{fig:boxplot} were selected in more than three out of the five cross-validation runs.
Although the GLCM exhibited significant differences across all statistics, 
it was confirmed that this was not selected for any of the five cross-validation runs.
In contrast, although the entropy of FOS and the mean, variance, skewness, 
and the kurtosis of ADF did not show significant differences in Figure~\ref{fig:boxplot}, 
they were selected at least once during the cross-validation.
This can be attributed to the stability of the FS algorithm.
Kalousis et al.~\cite{kalousis2007stability} defined the stability of a FS method as 
``the robustness of the feature preferences it produces to differences 
in training sets drawn from the same generating distribution.''
Dernoncourt et al.~\cite{dernoncourt2014analysis} reported that in the case of small sample size data, 
the stability of the FS method decreases, resulting in a higher likelihood of selecting irrelevant features.
In addition, they demonstrated that even with appropriate FS methods, 
the probability of selecting optimal features remains significantly low.
Therefore, to achieve higher generalization performance and more reliable factor analysis, 
it is essential to address the stability of the FS methods.
One possible approach would be to utilize domain knowledge~\cite{groves2013using}. 
This approach involves preserving potentially interesting relationships while excluding FS configurations that contradict domain knowledge.
Thus, irrelevant features may be excluded, potentially improving the stability of FS.

%%%%%%%%%  表：選択されたテクスチャ特徴量  %%%%%%%%%%
\begin{table*}[t]
  \centering
  
  \begin{threeparttable}[h] % 注釈
  \caption{Percentage of features selected through feature selection and dimensional compression in SVM with linear kernel using five-fold stratified cross-validation (training--test split) for feature selection.}
  \label{table:selected texture features}

  \begin{tabular*}{\textwidth}{@{\extracolsep{\fill}}lcccccc}
    \hline
    \multirow{2}{*}{Statistical measures$^{1}$} & \multicolumn{6}{c}{Texture analysis methods$^{2,\:3}$} \\
                     & FOS & GLDS & GLCM & GLRLM & ADF & RDF \\  
    \hline
    Mean             & *$\mathbf{1.0}$ & *$0.0$          & \textemdash{} & \textemdash{} & $0.2$           & *$\mathbf{0.8}$ \\
    Contrast         & *$\mathbf{1.0}$ & *$0.0$          & *$0.0$        & \textemdash{} & $0.0$           & *$\mathbf{1.0}$ \\
    Variance         & $0.0$           & *$0.0$          & \textemdash{} & \textemdash{} & $0.2$           & *$\mathbf{1.0}$ \\
    Skewness         & *$\mathbf{1.0}$ & *$\mathbf{0.8}$ & \textemdash{} & \textemdash{} & $0.2$           & *$\mathbf{0.8}$ \\
    Kurtosis         & *$\mathbf{1.0}$ & *$\mathbf{0.8}$ & \textemdash{} & \textemdash{} & $\mathbf{0.8}$  & *$\mathbf{1.0}$ \\
    Energy           & *$\mathbf{0.8}$ & *$0.2$          & \textemdash{} & \textemdash{} & *$\mathbf{1.0}$ & *$0.2$ \\
    Entropy          & $0.2$           & *$0.0$          & \textemdash{} & \textemdash{} & *$\mathbf{0.8}$ & *$\mathbf{0.6}$ \\
    \hdashline
    Correlation      & \textemdash{} & \textemdash{} & *$0.0$ & \textemdash{} & \textemdash{} & \textemdash{} \\
    Joint energy     & \textemdash{} & \textemdash{} & *$0.0$ & \textemdash{} & \textemdash{} & \textemdash{} \\
    Joint entropy    & \textemdash{} & \textemdash{} & *$0.0$ & \textemdash{} & \textemdash{} & \textemdash{} \\
    IDM              & \textemdash{} & \textemdash{} & *$0.0$ & \textemdash{} & \textemdash{} & \textemdash{} \\
    Inverse variance & \textemdash{} & \textemdash{} & *$0.0$ & \textemdash{} & \textemdash{} & \textemdash{}  \\
    \hdashline
    SRE              & \textemdash{} & \textemdash{} & \textemdash{} & *$\mathbf{0.6}$ & \textemdash{} & \textemdash{}  \\
    LRE              & \textemdash{} & \textemdash{} & \textemdash{} & *$0.2$          & \textemdash{} & \textemdash{}  \\
    GLN              & \textemdash{} & \textemdash{} & \textemdash{} & *$\mathbf{0.8}$ & \textemdash{} & \textemdash{}  \\
    RLN              & \textemdash{} & \textemdash{} & \textemdash{} & *$\mathbf{0.8}$ & \textemdash{} & \textemdash{}  \\
    RP               & \textemdash{} & \textemdash{} & \textemdash{} & *$\mathbf{0.6}$ & \textemdash{} & \textemdash{}  \\
    \hline
  \end{tabular*}

  \footnotesize{
  \begin{tablenotes}
    \item[$1$] IDM stands for inverse difference moment.
    SRE stands for short run emphasis.
    LRE stands for long run emphasis.
    GLN stands for gray level non-uniformity.
    RLN stands for run length non-uniformity.
    RP stands for run percentage.
    \item[$2$] FOS stands for first-order statistics.
    GLDS stands for gray level difference statistics.
    GLCM stands for gray level co-occurrence matrix.
    GLRLM stands for gray level run length matrix.
    ADF stands for angular distribution function via fourier power spectrum.
    RDF stands for radial distribution function via fourier power spectrum.
    \item[$3$] \textbf{Bold text} indicates that a feature was selected in more than three of the iterations during the five-fold stratified cross-validation.
    The asterisk (*) indicates that the feature is a characteristic for which a significant difference was confirmed through statistical hypothesis testing in Section~\ref{subsec: result of texture features}.
  \end{tablenotes}
  }
  
  \end{threeparttable}
\end{table*}
%%%%%%%%%%%%%%%%%%%%%%%%%%%%%%%%%%%%

\section{Conclusions} \label{sec:Conclusions}
The goal of this study was to develop a pathological diagnostic model for cardiomyopathy with high generalization performance, 
even in environments where collecting diverse and large sample sizes of endomyocardial biopsy specimens is difficult.
To achieve this, the effectiveness of texture features in the pathological diagnosis of cardiomyopathy was examined, 
and a model design suitable for small sample size data was assessed.

Regarding texture features, although several features clearly distinguished all three classes, 
many features were observed to distinctly separate one class from the others, as demonstrated qualitatively.
Additionally, the Kruskal-Wallis test and Benjamini-Hochberg method indicated that 32 out of the 39 texture features were useful for classification.
Regarding the model design, whether using classification models or dimensionality reduction methods, 
good predictive performance was achieved for the pathological diagnosis of the three myocardial states.
However, when applying DC alone, a discrepancy of more than 0.2 points between the training and test set evaluation values 
was observed, indicating an overfitting tendency.
In addition, when applying FS alone, only a single case of improvement in the generalization performance was observed 
on the validation or test sets compared to models without dimensionality reduction.
Conversely, the combination of FS and DC demonstrated the best generalization performance for the test set, 
outperforming both models without dimensionality reduction and those with single-dimensionality reduction methods.
These trends were consistent across all the classification models used in this study.

These experimental results suggest that even when the ratio of features to sample size is high, 
most features remain useful for classification, and the application of nested cross-validation and hyperparameter optimization, 
which are robust to small sample size data, may mitigate overfitting and enable high predictive performance.
Moreover, if most features are useful and FS does not substantially reduce the feature space, 
applying FS does not contribute significantly to improving generalization performance.
Furthermore, when the ratio of features to sample size is high, a multistage dimensionality reduction process may prove effective.
This process is likely to be effective, regardless of whether the decision boundary is 
linear, curvilinear, or vertical/horizontal to the axes, provided that the complexity of the model is controlled.
When these conditions were satisfied, the generalization performance of SVM with LK was the highest, 
achieving a macro-average F1-score of 0.949 in the test set.

These findings are expected to help solve various issues related to the shortage of pathology experts 
in cardiology by developing a pathology diagnostic model for cardiomyopathy with a high generalization performance.
Furthermore, the model design may be applicable to other diseases with insufficient data sizes, 
potentially contributing to the rapid adoption of ML models in medical practice.

This study has two main limitations.
First, it does not consider individual differences inherent in medical data.
In general, medical data contain several unobservable individual differences, such as lifestyle factors, 
genetic predisposition, environmental influences, psychological aspects, and socioeconomic conditions, 
which are known to degrade the generalization performance~\cite{ding2024deep}.
Therefore, it is essential to consider the effects of these individual differences 
when applying dimensionality reduction or hyperparameter optimization.
However, in this study, only two subjects were included per case, making it impossible to account for individual differences.
Thus, it is necessary to investigate the extent to which texture features and model designs evaluated 
in this study are effective for unseen individual differences.
Secondly, the need to explore methods more specifically tailored to small sample size data should be considered.
Although this study achieved a high generalization performance with a macro-average F1-score of 0.949 on the test set, 
certain aspects remain unaddressed.
Specifically, the stability of the FS methods 
has not yet been examined~\cite{kalousis2007stability, dernoncourt2014analysis}, 
and the application of dimensionality reduction methods tailored to 
high-dimensional low-sample size data have not yet been explored~\cite{yang2021improving, friedman1989regularized}.
Addressing these aspects is indispensable for achieving further improvements in generalization performance and effectiveness.
Therefore, these issues remain important subjects for future research.

\section*{Acknowledgments}
This work was supported in part by the 
Japan Society for the Promotion of Science (JSPS) Grants-in-Aid for Scientific Research (C) (Grant Nos. 23K11310 and 21K04535) and 
Nihon University Research Grant for 2023 [23-14].

\section*{Conflict of interest}
The authors declare that they have no known competing financial interests or personal relationships 
that could have appeared to influence the work reported in this paper.




\bibliographystyle{elsarticle-num}
%\bibliography{references.bib}

\end{document}